\documentclass[preprint,review,12pt]{elsarticle}
\usepackage{amssymb}
\usepackage{subfigure}
\usepackage{float}
\usepackage{booktabs}
\usepackage{amsmath}
\usepackage{array}

\usepackage[pagebackref=false,breaklinks=true,letterpaper=true,colorlinks,bookmarks=true]{hyperref}
\usepackage{color}
\usepackage{algorithm}
\usepackage{algorithmic}
\graphicspath{{figures/}}

\journal{Signal Processing: Image Communication}

\begin{document}

\begin{frontmatter}

\title{Kernel Estimation from Salient Structure for Robust Motion Deblurring\tnoteref{t1}}
%\tnoteref{t1}
\tnotetext[t1]{The MATLAB codes are now available at \textcolor[rgb]{1.00,0.00,0.00}{\url{https://www.dropbox.com/s/eixi8a2nsg15mhk/Deblurring_code_v2.zip}}}
\author[label1]{Jinshan Pan}
\ead{jspan@mail.dlut.edu.cn}
\author[label1,label2]{Risheng Liu}
\ead{rsliu@dlut.edu.cn}
\author[label1]{Zhixun Su\corref{cor1}}
\ead{zxsu@dlut.edu.cn}
\author[label3]{Xianfeng Gu}
\ead{gu@cs.sunysb.edu}
\cortext[cor1]{Corresponding author. Tel.: +86-411 84708351-8020.}
\address[label1]{School of Mathematical
Sciences, Dalian University of Technology, Dalian, China}
\address[label2]{Department of Biomedical Engineering, Faculty of Electronic Information and Electrical Engineering,
 Dalian University of Technology, Dalian, China}
\address[label3]{Department of Computer Science, Stony Brook University, Stony Brook, USA}

\begin{abstract}
Blind image deblurring algorithms have been improving steadily in the past years.
Most state-of-the-art algorithms, however, still cannot perform perfectly in challenging cases,
especially in large blur setting.
In this paper, we focus on how to estimate a good blur kernel from a single blurred image
based on the image structure.
We found that image details caused by blur could
adversely affect the kernel estimation, especially when the blur
kernel is large. One effective way to remove these details is to
apply image denoising model based on the Total Variation (TV).
First, we developed a novel method for computing image structures
based on TV model, such that the structures
undermining the kernel estimation will be removed.
Second, we applied a gradient selection method to mitigate the possible adverse
effect of salient edges and improve the robustness of kernel estimation.
Third, we proposed a novel kernel estimation method, which
is capable of removing noise and preserving the continuity in the kernel.
Finally, we
developed an adaptive weighted spatial prior to
preserve sharp edges in latent image restoration.
Extensive experiments testify to the effectiveness of our method on various
kinds of challenging examples.

\end{abstract}


\end{frontmatter}

% \linenumbers

\section{Introduction}
\label{sec:introduction}
Blind image deblurring is a challenging problem which has
drawn a lot of attention in recent years due to its
involvement of many challenges in problem formulation, regularization, and
optimization.
The formation process of motion blur is usually modeled as
\begin{equation}
\label{blur_model}
B = k\ast I + \varepsilon,
\end{equation}
where $B$, $I$, $k$ and $\varepsilon$ represent the blurred image, latent image,
blur kernel (a.k.a. point spread function, PSF) and the additive noise, respectively.
$\ast$ denotes the convolution operator. It is a well-known ill-posed inverse problem,
which requires regularization to alleviate its ill-posedness and stabilize the solution.

Recently, significant processes have been made in~\cite{Cho/et/al,Shan/et/al,Xu/et/al,Fergus/et/al,Joshi/et/al,Krishnan/CVPR2011,Levin/CVPR2011}.
The success of these methods comes from two important aspects: the sharp edge restoration
and noise suppression in smooth regions, which enable accurate kernel estimation.

\begin{figure}[!t]
\centering
\includegraphics[angle=0, width=0.95\textwidth]{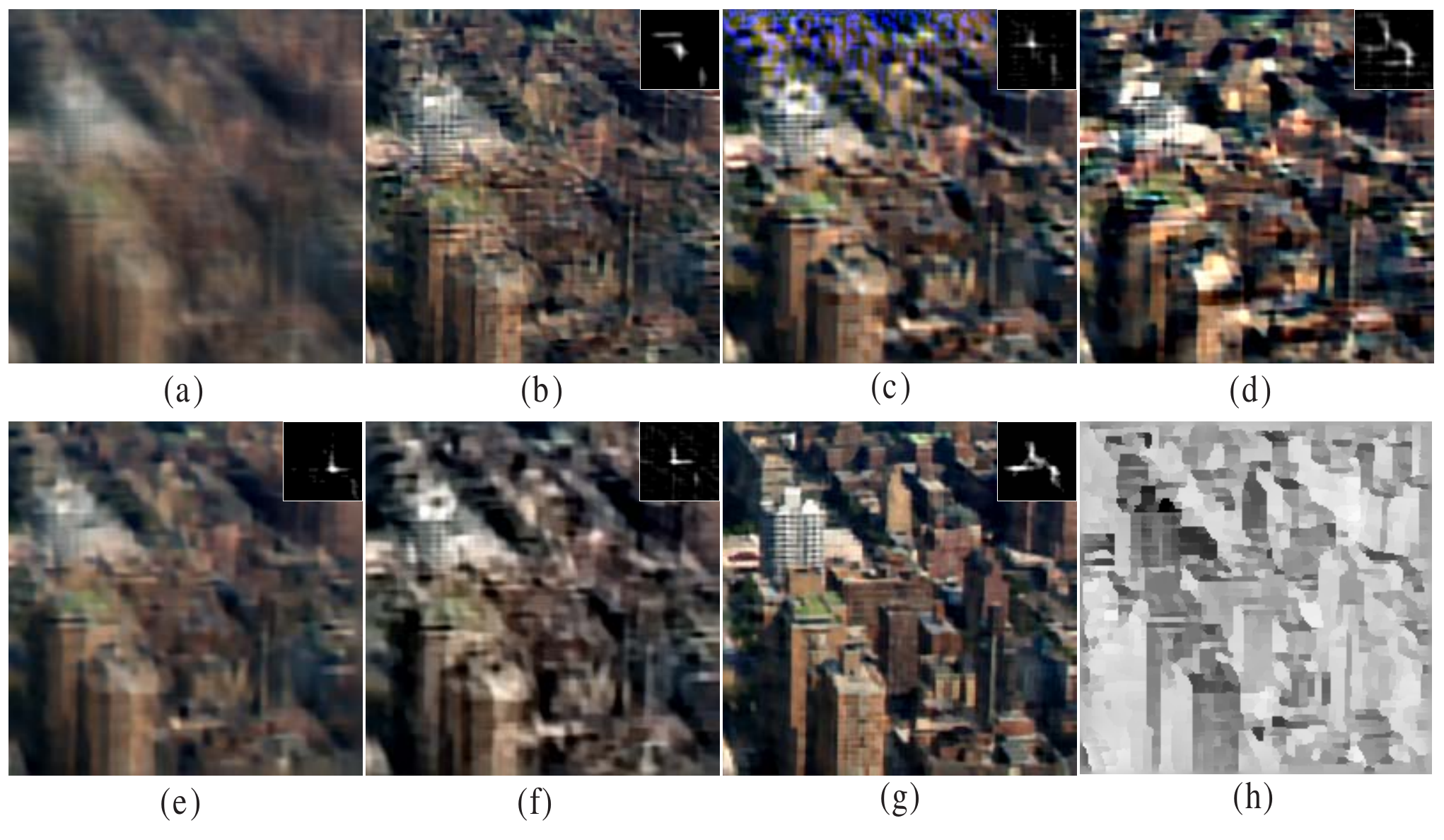}
\caption{A challenging example.
(a) Blurred image. (b) Result of Shan {\it et al.}~\cite{Shan/et/al}. (c) Result of Cho and Lee~\cite{Cho/et/al}.
(d) Result of Xu and Jia~\cite{Xu/et/al}. (e) Result of Levin {\it et al.}~\cite{Levin/CVPR2011}.
(f) Result of Krishnan {\it et al.}~\cite{Krishnan/CVPR2011}. (g) Our result.
(h) Our final salient edges $\nabla S$ (detailed further below) visualized by using Poisson reconstruction method.
The size of motion blur kernel is $45\times 45$.}
\label{fig:details result}
\end{figure}

However, blurred image with some complex structures or large blur will fail
most of state-of-the-art blind deblurring methods.
Taking Fig.~\ref{fig:details result}(a) as an example,
the motion blur is very large due to the camera shake.
In addition, the blurred image also contains complex structures.
As shown in Figs.~\ref{fig:details result}(b) - (f), some state-of-the-art
methods~\cite{Cho/et/al,Shan/et/al,Xu/et/al,Krishnan/CVPR2011,Levin/CVPR2011}
have difficulty in restoring or selecting useful sharp edges for kernel estimation due to the large blur and complex structures.
Thus, the correct blur kernels are not obtained. This inevitably makes the final deblurred results unreliable.

We address this issue and propose a new kernel estimation method based on the reliable structures.
Our method is a new selection scheme to select reliable structures according to the image characteristics.
Thus, it is able to get useful sharp edges for kernel estimation.
Our deblurred result shown in Fig.~\ref{fig:details result}(g) contains fine textures,
and the kernel estimate is also better than others.

In addition, noisy kernels also damage the kernel estimation,
which further leads to unreliable deblurred results.
Therefore, removing noise in the kernels is also very important in kernel estimation.

Based on above analysis, we develop several strategies which are significantly different
from previous works in the following aspects.
\begin{enumerate}
    \item First, to remove detrimental structures and obtain useful information for kernel estimation,
    we develop a novel adaptive structure selection method which can choose reliable structures effectively.
    \item To preserve the sparsity and continuity of blur kernels, we propose a new robust kernel
    estimation method by introducing a powerful spatial prior, which also helps remove the noise effectively.
    \item Finally, we introduce a simple adaptive regularization term that combines the final
    salient structures to guide the latent image restoration, which is able to preserve sharp edges
    in the restored image.
\end{enumerate}

We apply our method to some challenging examples, such
as images with complex tiny structures or with large blur
kernels, and verify that it is able to provide reliable kernel
estimates which are noiseless and also satisfy the sparsity and
continuity well. Moreover, high-quality final restored images
can be obtained.

%-------------------------------------------------------------------------
\section{Related Work}
\label{Related Work}
Image deblurring is a hot topic in image processing and
computer vision communities.
In single image blind deblurring, early approaches usually imposed constraints on motion
blur kernel and used parameterized forms for the kernels~\cite{Chen/et/al,Chan/and/Wong}.
Recently, Fergus {\it et al.}~\cite{Fergus/et/al} adopted a zero-mean Mixture of
Gaussian to fit for natural image gradients. A variational
Bayesian method was employed to deblur an image.
Shan {\it et al.}~\cite{Shan/et/al} used a certain parametric
model to approximate the heavy-tailed natural image prior.
Cai {\it et al.}~\cite{Cai/cvpr09} assumed that the latent images and kernels can
be sparsely represented by an over-complete dictionary
and introduced a framelet and curvelet system to obtain the sparse representation for images and kernels.
Levin {\it et al.}~\cite{Levin/CVPR2009} illustrated the limitation of the
simple maximum a posteriori (MAP) approach,
and proposed an efficient marginal likelihood
approximation in~\cite{Levin/CVPR2011}.
Krishnan {\it et al.}~\cite{Krishnan/CVPR2011} introduced a new
normalized sparsity prior to estimate blur kernels.
Goldstein and Fattal~\cite{Goldstein/eccv2012} estimated blur kernels by spectral irregularities.
However, the kernel estimates of the aforementioned works usually contain some noise.
The hard thresholding to the kernel elements method will destroy the inherent structure of kernels.

Another group methods~\cite{Cho/et/al,Xu/et/al,Joshi/et/al,Joshi/phd,radon/cvpr/ChoPHF11}
employed an explicit edge prediction step for kernel estimation.
In~\cite{Joshi/et/al}, Joshi {\it et al.} computed sharp edges by first locating step edges
and then propagating the local intensity extrema towards the edge.
Cho {\it et al.}~\cite{radon/cvpr/ChoPHF11} detected sharp edges from blurred images
directly and then the Radon transform was employed to estimate the blur kernel.
However, these methods have difficulty in dealing with large blur.
Cho and Lee~\cite{Cho/et/al} used bilateral filtering together with shock
filtering to predict sharp edges iteratively and then selected the salient edges for kernel estimation.
However, the Gaussian priors used in this method can not keep the sparsity of the motion blur kernel and
the image structure. The final result usually contains noise.
Xu and Jia~\cite{Xu/et/al} proposed an effective mask computation algorithm to adaptively
select useful edges for kernel estimation. The kernel refinement was achieved by using
iterative support detection (ISD) method~\cite{ISD}.
However, this method ignores the continuity of the motion blur kernel.
The estimated kernels contain some noise occasionally.
Hu and Yang~\cite{hu/eccv12/region} learned good regions for kernel estimation and employed
method~\cite{Cho/et/al} to estimate kernels.
Although the performance is greatly improved,
the sparsity and continuity of blur kernels still can not be guaranteed.

After obtaining the blur kernel, the blind deblurring problem
becomes a non-blind deconvolution.
Early approaches such as Wiener filter and Richardson-Lucy
deconvolution~\cite{Lucy/deconvolution} usually suffer from noise and ringing artifacts.
Yuan {\it et al.}~\cite{Yuan/non/blind/deblurring/tog08} proposed a progressive inter-scale and intra-scale
based on the bilateral Richardson-Lucy method to reduce ringing artifacts.
Recent works mainly focus on the natural image statistics~\cite{Shan/et/al,Levin/et/al} to
keep the properties of latent images and suppress ringing artifacts.
Joshi {\it et al.}~\cite{joshi/color/prior/cvpr09}
used local color statistics derived from the image as a constraint to
guide the latent images restoration.
The works in~\cite{Xu/et/al,Wang/Yang/Yin/Zhang} used TV regularization to restore latent images,
but the isotropic TV regularization will result in stair-casing effect.

It is noted that there also have been active researches on spatially-varying
blind deblurring methods. Interested readers are referred
to~\cite{Whyte/cvpr10,non/uniform/deblur/joshi,Gupta/non/uniform/eccv10,
hirsch/iccv11/non/uniform/deblurring,hui/ji/cvpr12/non/uniform/deblurring}
for more details.

%-------------------------------------------------------------------------
%\section{Motion Deblurring from Salient Structure}
\section{Kernel Estimation from Salient Structure}
\begin{figure}[!t]
\begin{center}
\includegraphics[angle=0, width=0.95\textwidth]{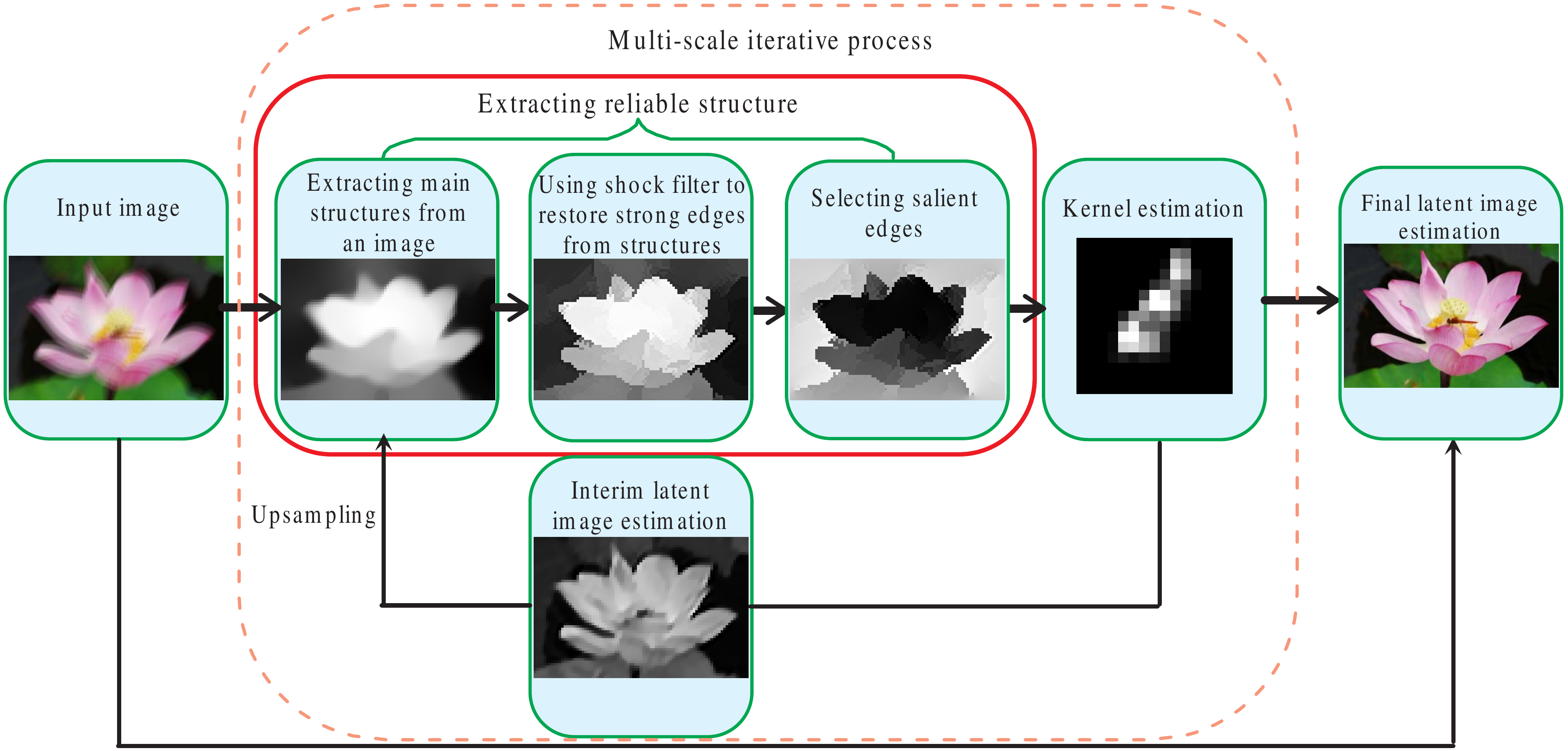}
\end{center}
   \caption{The flowchart of our algorithm. The dotted line box encloses the
process of kernel estimation.}
\label{fig:short}
\end{figure}

We find that different extraction of structure leads to different deblurred results,
and extracting reliable structure is critical to deblurring.
Thus, we focus on extracting more reliable structures, which is achieved by several key steps.
First, we extract the main image structure (the first part of the solid red box in Fig.~\ref{fig:short}).
Then, a shock filter is applied to get the enhanced structure
(the second part of the solid red box in Fig.~\ref{fig:short}).
Finally, some salient edges with large pixel values will be selected for the kernel
estimation (the third part of the solid red box in Fig.~\ref{fig:short}).
The details of this process will be
discussed in Section~\ref{subsec: comppute structure},
and the corresponding reasoning will be provided in Section~\ref{sssec: Analysis on Structure Selection Method}.

%%%%%%%%%%%%%%%%%%%%%%%%%%%%%%%%%%%%%%%%%%%%%%%%%%%%%%%%
It is noted that noisy interim kernels will also
damage the interim latent image estimation,
which further leads to unreliable kernels during the kernel refinement.
We propose a robust kernel estimation method
which combines the gradient properties of kernels to overcome this problem.
Detailed analysis will be discussed in Section~\ref{ssec: Kernel Estimation}.
The dotted line box shown in Fig.~\ref{fig:short} encloses the process of kernel estimation in details.
%----------------------------------------------------------------------------------------------------

\subsection{Extracting Reliable Structure}
\label{subsec: comppute structure}
\begin{figure}[!t]
\setlength{\abovecaptionskip}{-0.2cm} %
\begin{center}
\includegraphics[angle=0, width=0.95\textwidth]{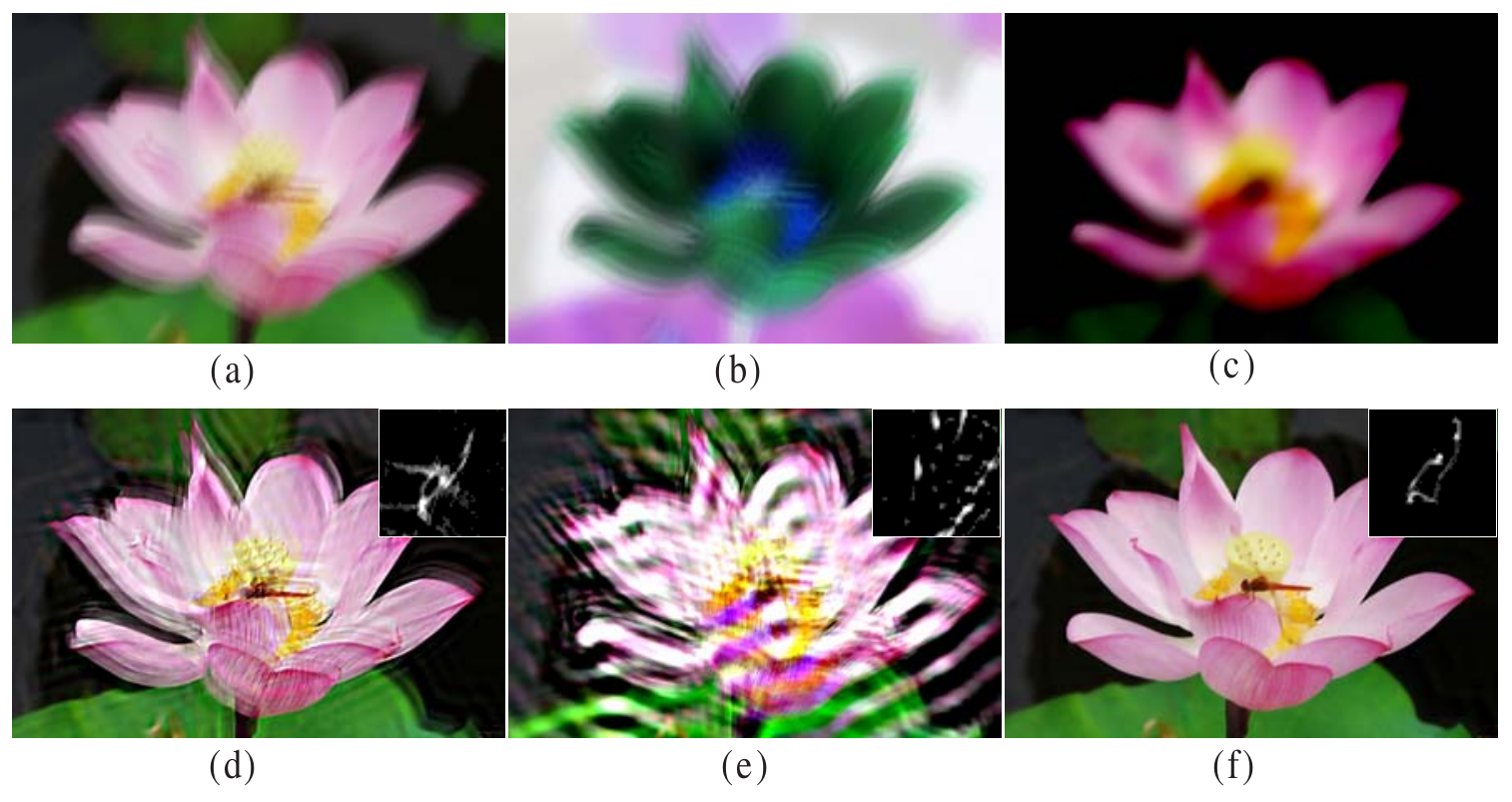}
\end{center}
   \caption{Different structures leading to different deblurred results.
   (a) Blurred image.
   (b) Texture component $I_T$.
   (c) Structure component $I_s$.
   (d) Results without performing model~(\ref{eq:TV structure}).
   (e) Results by using $I_T$ in the process of kernel estimation.
   (f) Results by using $I_s$ in the process of kernel estimation.}
\label{fig:structure of reslut}
\end{figure}

Our method for adaptively selecting salient edges mainly relies on
the idea of structure-texture decomposition method~\cite{Rudin/Osher/Fatemi}.
For an image $I$ with pixel intensity value $I(\textbf{x})$, the structure
part is given by the optimizer of the following energy:
\begin{eqnarray}\label{eq:TV structure}
\min_{I_s}\sum_{\textbf{x}}\|\nabla I_s(\textbf{x})\|_2+\frac{1}{2\theta}(I_s(\textbf{x})-I(\textbf{x}))^2,
\end{eqnarray}
where $\theta$ is an adjustable parameter.
The image $I$ is decomposed into the structure component $I_s$ (shown in Fig.~\ref{fig:structure of reslut}(c))
and the texture component $I_T =I-I_s$ (shown in Fig.~\ref{fig:structure of reslut}(b)).
The structure component $I_s$ contains the major objects in
the image while $I_T$ includes fine-scale details and noise.

Fig.~\ref{fig:structure of reslut}(f) demonstrates that the
accuracy of kernel estimate is greatly improved by performing model~(\ref{eq:TV structure}).
However, model~(\ref{eq:TV structure}) may lead to stair-casing effect
in smooth area. This will cause gradient distortion and introduce
inaccuracy for kernel estimation. A simple way to mitigate this
effect is to adjust the value of $\theta$ to be large in the smooth areas,
and small near the edges.
To that end,
we adopt the following adaptive model to select the
main structure of an image $I$:
\begin{eqnarray}\label{eq:improved TV model}
\min_{I_s}\sum_{\textbf{x}}\|\nabla I_s(\textbf{x})\|_2+\frac{1}{2\theta \omega(\textbf{x})}(I_s(\textbf{x})-I(\textbf{x}))^2,
\end{eqnarray}
where $\omega(\textbf{x})=\exp(-|r(\textbf{x})|^{0.8})$ and
$r(\textbf{x})$ is defined as
\begin{align}\label{eq:r-map}
r(\textbf{x})=\frac{\|\sum_{\textbf{y}\in N_h(\textbf{x})}\nabla B(\textbf{y})\|_2}{\sum_{\textbf{y}\in N_h(\textbf{x})}\|\nabla B(\textbf{y})\|_2 + 0.5},
\end{align}
in which $B$ is the blurred image, and $N_h(\textbf{x})$ is an $h \times h$ window
centered at pixel $\textbf{x}$. A small $r$ implies that the local region is flat,
whereas large $r$ implies existing strong image structures in the
local window. This equation is first employed by~\cite{Xu/et/al} to remove
some narrow strips that may undermine the kernel estimation.
However, model~(\ref{eq:improved TV model}) keeps the similar advantages
to those of~\cite{Xu/et/al} due to the adaptive weight $\omega(\textbf{x})$.
It also has a strong penalty to these areas which are flat or contain
narrow strips as well.
\begin{figure}[!t]
\setlength{\abovecaptionskip}{-0.3cm} %
\setlength{\belowcaptionskip}{-0.3cm} %
\begin{center}
\includegraphics[angle=0, width=0.98\textwidth]{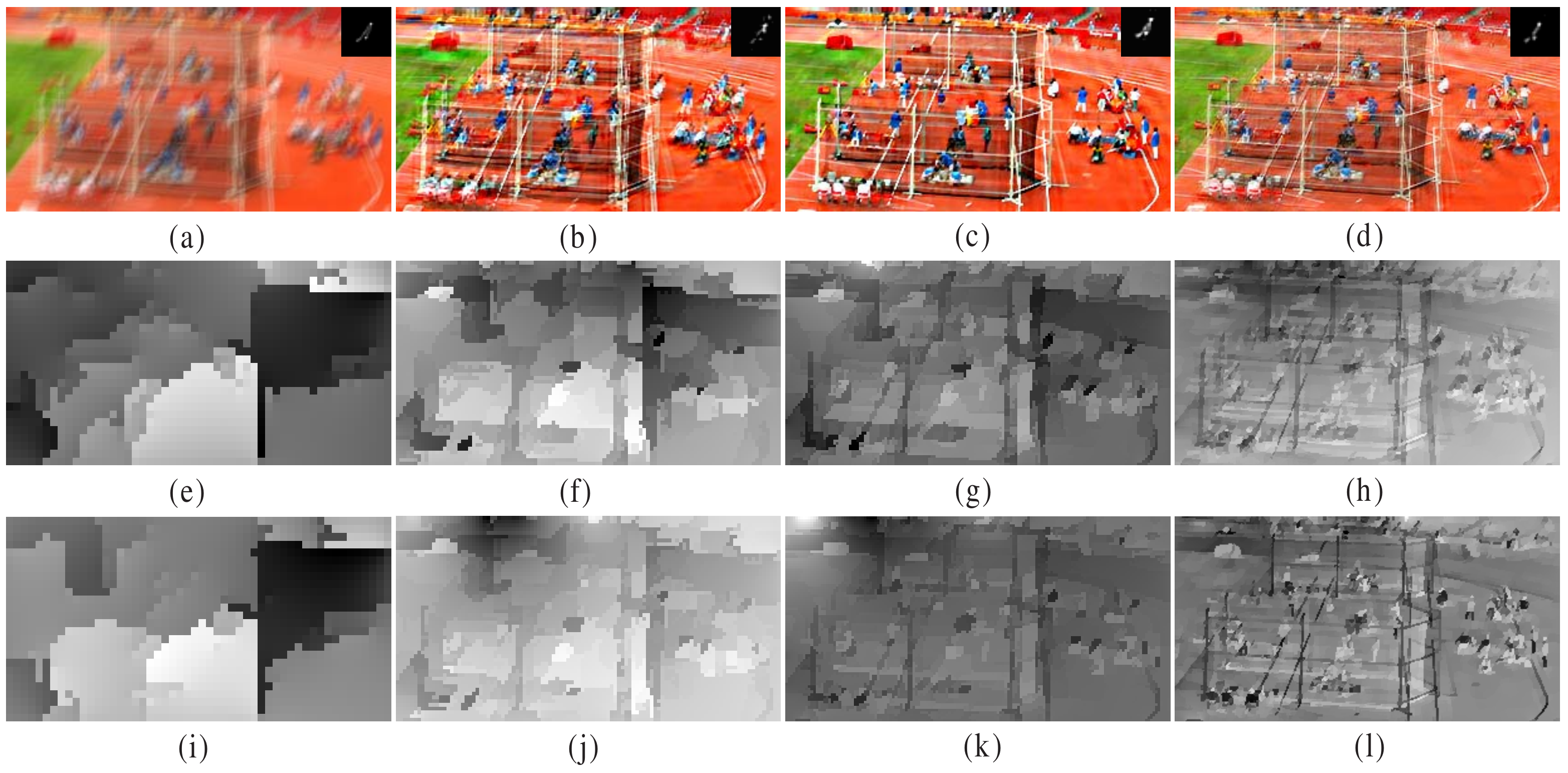}
\end{center}
\caption{Comparison of results with models~(\ref{eq:TV structure}) and~(\ref{eq:improved TV model}).
   (a) Blurred image and truth kernel.
   (b) Results with model~(\ref{eq:TV structure}).
   (c) Results with model~(\ref{eq:improved TV model}).
   (d) Results of~\cite{Xu/et/al}.
   (e) - (h) Interim $\nabla S$ maps with model~(\ref{eq:TV structure}).
   (i) - (l) Interim $\nabla S$ maps with model~(\ref{eq:improved TV model}).
   The final results (including the kernel estimate and deblurred result)
   shown in (c) performs better than others.}
\label{fig:compare TV}
\end{figure}

To demonstrate the validity of model~(\ref{eq:improved TV model}),
we conduct an experiment shown in Fig.~\ref{fig:compare TV}.
As shown in Fig.~\ref{fig:compare TV}(a), the blurred image
contains some complex structures, which may have detrimental effects on kernel estimation.
Due to adopting the adaptive weight $\omega(\textbf{x})$ in model~(\ref{eq:improved TV model}),
kernel estimation result by model~(\ref{eq:improved TV model}) is significantly better than that by
model~(\ref{eq:TV structure}).
Furthermore, compared with $\nabla S$ (detailed further below) maps shown in Figs.~\ref{fig:compare TV} (e) - (h)
and Figs.~\ref{fig:compare TV} (i) - (l),
%the adaptive structure selection model~(\ref{eq:improved TV model}) can
%not only remove tiny details and noise but also retain some useful structures for kernel estimation.
both models~(\ref{eq:TV structure}) and~(\ref{eq:improved TV model}) are able to
select main structures of an image, but model~(\ref{eq:improved TV model}) can retain some useful structures
for kernel estimation.

After computing $I_s$,
we compute the
enhanced structure $\tilde{I_s}$ by a shock filter~\cite{Shock/filter}:
%\begin{eqnarray}\label{eq:shock filter}
%%\tilde{I_s}=-\text{sign}(\triangle I_s)\| \nabla I_s \|_2,
%\partial \tilde{I_s}/\partial t = -\text{sign}(\triangle I_s)\| \nabla I_s \|_2,
%\end{eqnarray}

\begin{equation}
\label{eq:shock filter}
\begin{aligned}
&\partial \tilde{I_s}/\partial t = -\text{sign}(\triangle \tilde{I_s})\| \nabla \tilde{I_s} \|_2,\\
&\tilde{I_s}|_{t=0} = I_s,
\end{aligned}
\end{equation}

where $\triangle I=I_x^2I_{xx}+2I_xI_yI_{xy}+I_y^2I_{yy}$.

%%%%%%%%%%%%%%%%%%%%%%%%%%%%%%%%%%%%%%%%%%%%%%%%%%%%%%%%%%%%%%%%%%%%%%%%%
%%%%%%%%%%%%%%%%%%%%%%%%%%%%%%%%%%%%%%%%%%%%%%%%%%%%%%%%%%%%%%%%%%%%%%%%%
Finally, we compute salient edges $\nabla S$ which will be used
to guide the kernel estimation:
\begin{eqnarray}\label{eq:shock structure model}
\nabla S = \nabla \tilde{I_{s}}H(\textbf{G}, t),
\end{eqnarray}
where $H(\textbf{G}, t)$ is the unit binary mask function which is defined as
\begin{eqnarray}\label{eq:Mask function }
H(\textbf{G},t)=\left\{\begin{array}{ll}1, & \mbox{}\ G_i\geqslant t,\\0, & \mbox{}\ $otherwise$,\end{array}\right.
\end{eqnarray}
and $\textbf{G}=(\| \nabla \tilde{I_{s}} \|_2,\| \partial_x \tilde{I_{s}}\|_1/5\sqrt{2},\| \partial_y\tilde{I_{s}}\|_1/5\sqrt{2})$.
The parameter $t$ is a threshold of the gradient magnitude $\| \nabla \tilde{I_{s}} \|_2$.
By applying Eq.~(\ref{eq:shock structure model}), some noise in the $\nabla \tilde{I_{s}}$ will be eliminated.
Thus, only the salient edges with large values have influences on the kernel estimation.
%%%%%%%%%%%%%%%%%%%%%%%%%%%%%%%%%%%%%%%%%%%%%%%%%%%%%%%%%%%%%%%%%%%%%%%%%
%%%%%%%%%%%%%%%%%%%%%%%%%%%%%%%%%%%%%%%%%%%%%%%%%%%%%%%%%%%%%%%%%%%%%%%%%

It is noted that kernel estimation will be unreliable
when a few salient edges are available for estimation.
To solve this problem, we adopt several strategies as follows.

First, we adaptively set the
initial values of $t$ according to the method of~\cite{Cho/et/al}
at the beginning of the iterative deblurring process.
Specifically, the directions of image gradients are initially quantized into four groups.
$t$ is set to guarantee that at least $\frac{1}{2}\sqrt{N_IN_k}$ pixels participate in
kernel estimation in each group, where $N_I$ and $N_k$ denote the total number of pixels in the input image
and the kernel, respectively.

Then,
as the iteration goes in the deblurring process,
we gradually decrease the values of $\theta$ and $t$ at each iteration to include more
edges for kernel estimation according to~\cite{Xu/et/al}.
This adaptive strategy can allow for inferring subtle structures during kernel refinement.

Figs.~\ref{fig:compare TV} (e) - (l) show some interim $\nabla S$ maps in the iterative deblurring process.
One can see that as the iteration goes, more and more sharp edges are included for kernel estimation.

%%%%%%%%%%%%%%%%%%%%%%%%%%%%%%%%%%%%%%%%%%%%%%%%%%%%%%%%%%%%%%%%%%%%%%%%%%%%%%%%%%%%%%%%%%%%%%%%%
%% 2013-02-21
\subsubsection{Analysis on Structure Selection Method}
\label{sssec: Analysis on Structure Selection Method}
To better understand our structure selection method, we use a 1D signal to
provide more insightful analysis on how our method can help kernel estimation.
\begin{figure}[!t]
\setlength{\abovecaptionskip}{-0.2cm} %缩小caption和图像之间的距离
\begin{center}
\includegraphics[angle=0, width=0.95\textwidth]{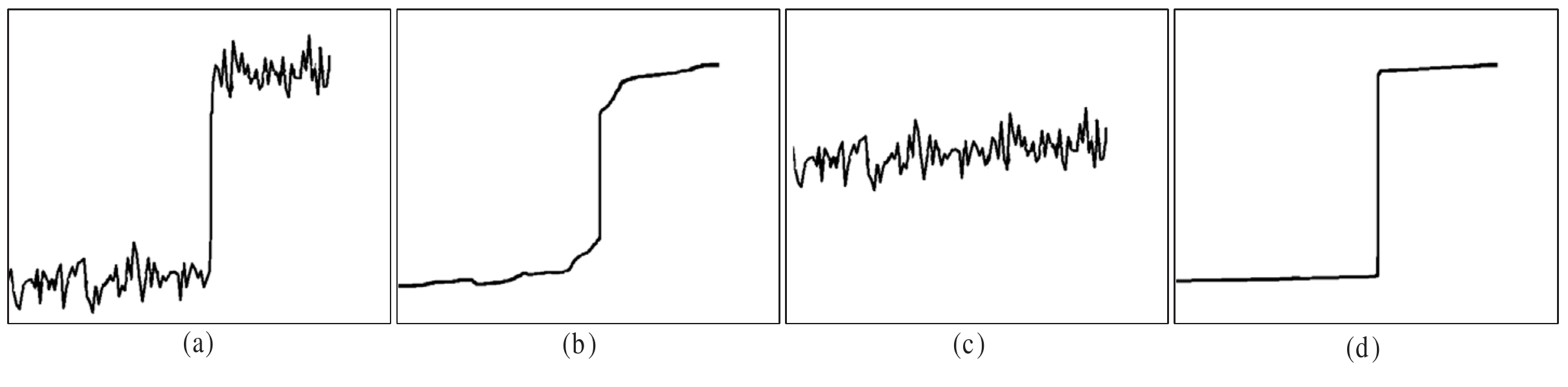}
\end{center}
   \caption{1D signal illustration.
   The 1D signal (a) is decomposed into two main components:
   (b) structure component and
   (c) texture component by using model~(\ref{eq:improved TV model}).
   (d) Sharp signal.
   Signals shown in this figure are obtained from an image scanline.}
\label{fig:signal-illustration}
\end{figure}

Given a signal (e.g., Fig.~\ref{fig:signal-illustration}(a)), we can decompose it  %% an arbitrary
into the structure component (Fig.~\ref{fig:signal-illustration}(b))
and texture component (Fig.~\ref{fig:signal-illustration}(c)) by using model~(\ref{eq:improved TV model}).
For the structure component (Fig.~\ref{fig:signal-illustration}(b)),
we can use a shock filter and Eq.~(\ref{eq:shock structure model}) to get
a sharp signal (Fig.~\ref{fig:signal-illustration}(d)) that is similar to the step signal.
Step signal, however, usually succeeds in the kernel estimation, which has been proved by
many previous works~\cite{Levin/CVPR2009,Joshi/et/al}.
In contrast, the texture component (Fig.~\ref{fig:signal-illustration}(c)) usually fails in the kernel estimation.
There are two mainly reasons: (1) The texture component contains noise which damages the kernel estimation;
(2) The size of texture component is relatively small. Blurring reduces its peak height - that is,
the shape of texture component is destroyed seriously after blur.
%%%%%%%%%%%%%%%%%%%%%%%%%%%%%%%%%%%%%%
Therefore, recovering the sharp version
of texture component from the blurred version %by a shock filter and Eq.~(\ref{eq:shock structure model})
is a very difficult problem.
%%%%%%%%%%%%%%%%%%%%%%%%%%%%%%%%%%%%%%
As a result, a correct kernel estimate is hard to be obtained by texture
component (e.g., Fig.~\ref{fig:structure of reslut}(e)).

More generally,
natural images can be regarded as 2D signals, which can be composed of many local
1D signals (Fig.~\ref{fig:signal-illustration}(a)).
This further demonstrates our method is valid.
%%%%%%%%%%%%%%%%%%%%%%%%%%%%%%%%%%%%%%%%%%%%%%%%%%%%%%%%%%%%%%%%%%%%%%%%%%%%%%%%%
We have also performed lots of experiments to verify the validity of our method.
The effectiveness of salient edges will be detailed in Section~\ref{sec: More Analysis on Kernel Estimation}.
%%%%%%%%%%%%%%%%%%%%%%%%%%%%%%%%%%%%%%%%%%%%%%%%%%%%%%%%%%%%%%%%%%%%%%%%%%%%%%%%%%%%%%%%%%%%%%%%%
%%%%%%%%%%%%%%%%%%%%%%%%%%%%%%%%%%%%%%%%%%%%%%%%%%%%%%%%%%%%%%%%%%%%%%%%%%%%%%%%%%%%%%%%%%%%%%%%%
%\subsection{Kernel Estimation from Salient Structure}
\subsection{Kernel Estimation}
\label{ssec: Kernel Estimation}
%%%%%%%%%%%%%%%%%%%%%%%%%%%%%%%%%%%%%%%%%%%%%%%%%%%%%%%
The motion blur kernel describes the path of camera shake
during the exposure. Most literatures assume that distributions
of blur kernels can be well modeled by a Hyper-Laplacian,
based on which the corresponding model for kernel estimation
is
\begin{eqnarray}\label{eq: final kernel estimation}
&&\min_k \| \nabla B-k*\nabla S\|_2^2+\gamma \|k\|_{\alpha}^{\alpha},\nonumber\\
&&\text{s.t.} \quad k(\textbf{x})\geq 0,\quad\sum_{\textbf{x}} k(\textbf{x}) = 1,
\end{eqnarray}
where $0 < \alpha \leqslant 1$.

\begin{figure}
\begin{center}
\includegraphics[angle=0, width=0.95\textwidth]{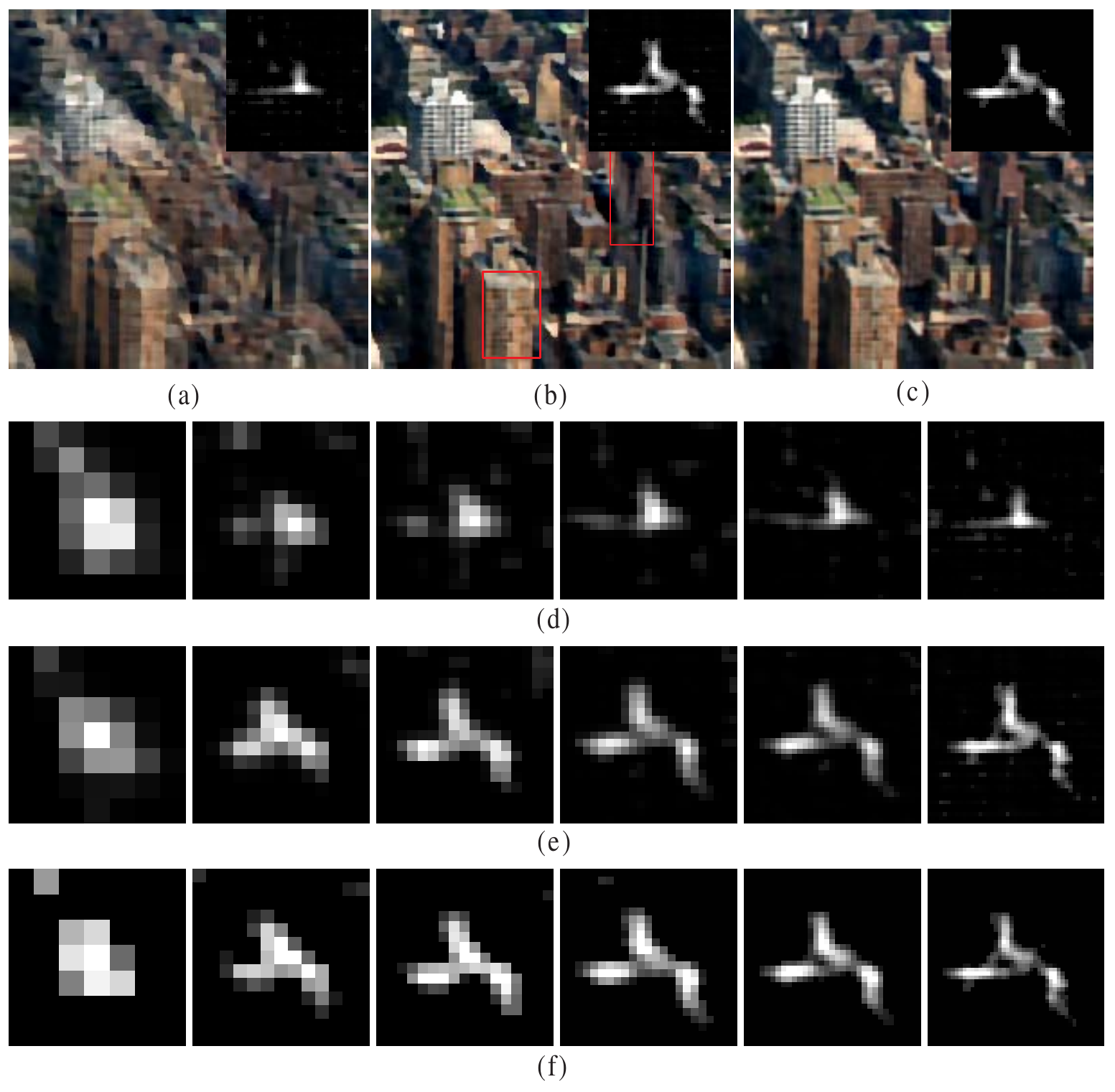} %
\end{center}
\caption{Comparison of results with different spatial priors of motion blur kernel.
In (a), the kernel estimate is obtained by using the kernel estimation model
that is employed by~\cite{radon/cvpr/ChoPHF11}.
The kernel estimate shown in (b) is obtained by using model~(\ref{eq: final kernel estimation}).
The kernel estimate shown in (c) is obtained by using model~(\ref{eq: improved final kernel estimation}).
(d), (e), and (f) show the iterations of kernel estimates by model in~\cite{radon/cvpr/ChoPHF11},
model~(\ref{eq: final kernel estimation}), and model~(\ref{eq: improved final kernel estimation}), respectively.
Deblurred result in (c) outperforms the result in (b) (e.g., the parts in the red boxes).
The results (c) and (f) of our method are the best.
}
\label{fig:PSF estimation with priors}
\end{figure}

Although model~(\ref{eq: final kernel
estimation}) can preserve the sparsity prior effectively, it
does not ensure the continuity of blur kernel and sometimes induces
noisy kernel estimates (e.g., the kernel estimate shown in Fig.~\ref{fig:PSF estimation with priors}(b)).
%%%%%%%%%%%%%%%%%%%%%%%%%%%%%%%%
Another critical problem is that the imperfectly estimated salient edges $\nabla S$ used
in model~(\ref{eq: final kernel estimation}) also lead to a noisy kernel estimate.
%%%%%%%%%%%%%%%%%%%%%%%%%%%%%%%%%%%%%%%%%%%%%
Figs.~\ref{fig:PSF estimation with priors}(a) and (b) show that
the correct deblurred results will not be obtained due to the
influence of noisy kernel estimates. From this example,
one can infer that noisy interim kernels will also damage
the interim latent image estimation which further may damage
the following estimated kernels during the kernel refinement.

%%%%%%%%%%%%%%%%%%%%%%%%%%%%%%%%%%%%%%%%%%%%%%%%%%%%%%%%%%%%%%%%%%%%%%%%%%%%%%%%%%%%
To overcome these problems, we constrain the gradients to preserve the continuity of kernel.
Considering the speciality of kernel, we introduce a new spatial term $\mathcal{C}(k)$,
which is defined as
\begin{eqnarray}\label{eq: constraint}
\mathcal{C}(k)=\#\{\textbf{x}| \quad |\partial_x k(\textbf{x})|+|\partial_y k(\textbf{x})|\neq 0\},
\end{eqnarray}
i.e., $\mathcal{C}(k)$ counts the number of pixels whose gradients are
non-zeros. It not only can keep the structure of kernel effectively but also remove some noise.

Based on the above considerations, our kernel estimation model is defined as
\begin{eqnarray}\label{eq: improved final kernel estimation}
&&\min_k\| \nabla B-k*\nabla S\|_2^2+\gamma \|k\|_{\alpha}^{\alpha}+\mu \mathcal{C}(k),\nonumber\\
&&\text{s.t.}\quad k(\textbf{x})\geq 0, \quad\sum_{\textbf{x}} k(\textbf{x}) = 1,
\end{eqnarray}
where the parameter $\mu$ controls the smoothness of $k$.
Model~(\ref{eq: improved final kernel estimation}) is robust to noise and can
preserve both sparsity and continuity of kernel. This is mainly because:
\begin{enumerate}
    \item The salient edges in the first term provide reliable edge information;
    \item The second term provides a sparsity prior for the kernel;
    \item The spatial term $\mathcal{C}(k)$ makes the kernel sparse and also discourages discontinuous points, hence promoting continuity.
\end{enumerate}
%%%%%%%%%%%%%%%%%%%%%%%%%%%%%%%%%%%%%%%%%%%%%%%%%%%%%%%%%%%%%%%%%%%%%%%%%%%%%%%%%%%%

Note that model~(\ref{eq: improved final kernel estimation}) is difficult to be minimized directly
as it involves a discrete counting metric. Similar to the strategy of~\cite{Chen/Jia},
we approximate it by alternately minimizing
\begin{eqnarray}\label{eq: L0 smooth_AM}
&&\min_{k}\| \nabla B-k*\nabla S\|_2^2 + \gamma \|k\|_{\alpha}^{\alpha},\nonumber\\
&&\text{s.t.}\quad k(\textbf{x})\geq 0, \quad\sum_{\textbf{x}} k(\textbf{x}) = 1,
\end{eqnarray}
and
\begin{eqnarray}\label{eq: L0 smooth_AM-Xu}
\min_{\hat{k}}\|\hat{k} - k\|_2^2 + \mu \mathcal{C}(\hat{k}).
\end{eqnarray}

Model~(\ref{eq: L0 smooth_AM}) can be optimized by using the constrained iterative
reweighed least square (IRLS) method~\cite{Levin/et/al}.
Specifically, we empirically run IRLS method for 3 iterations.
In the inner IRLS system, the optimization reduces to a quadratic programming problem (see the claim in~\cite{Levin/CVPR2011})
and the dual active-set method is employed to solve this quadratic programming.

For model~(\ref{eq: L0 smooth_AM-Xu}),
we employ the alternating optimization method in~\cite{Xu/L0/smooth}
to solve it.
Alg.~\ref{alg:2} illustrates the implementation details of model~(\ref{eq: improved final kernel estimation}).

%%%%%%%%%%%%%%%%%%%%%%%%%%%%%%%%%%%%%%%%%%%%%%%%%%%%%%%%%%%%%%%%%%%%%%%%%%%%%%%%%%%%%%%%%%
\vspace*{-0pt}
\begin{algorithm}\caption{Kernel Estimation Algorithm}
%\small
\label{alg:2}
\begin{algorithmic}
\STATE {\textbf{Input:} Blurred image $B$, salient edges $\nabla S$,
and the initial values of $k$ from previous iterations;}
\FOR {$i = 1 \to Itr$ ($Itr$: number of iterations)}
\STATE {Solve for $k$ by minimizing model~(\ref{eq: L0 smooth_AM});}
%\STATE {Normalize the kernel $k$ into the range [0,1];}
\STATE {Solve for $\hat{k}$ by minimizing model~(\ref{eq: L0 smooth_AM-Xu});}
\STATE {$k \gets \hat{k}$;}
\ENDFOR
\STATE \textbf{Output:} Blur kernel $k$.
\end{algorithmic}
\end{algorithm}
\vspace*{-0.2pt}
%%%%%%%%%%%%%%%%%%%%%%%%%%%%%%%%%%%%%%%%%%%%%%%%%%%%%%%%%%%%%%%%%%%%%%%%%%%%%%%%%%%%%%%%%%%

Here we empirically set $Itr = 2$ and $\alpha = 0.5$ in our experiments.
The parameter $\mu$ is chosen according to the size of kernels.
Fig.~\ref{fig:PSF estimation with priors}(f) shows the effectiveness of our model~(\ref{eq: improved final kernel estimation}).
From Fig.~\ref{fig:PSF estimation with priors}, one can see that although the structure selection method is the same,
the performances of kernel estimates by different models are different.
The estimated kernels by using the traditional gradient constraint
that is employed by~\cite{radon/cvpr/ChoPHF11} are unreliable (i.e., Fig.~\ref{fig:PSF estimation with priors}(d)).
As a result, these imperfect kernels further damage the following estimated results.
Fig.~\ref{fig:PSF estimation with priors}(e) shows that the kernel
estimates by using model~(\ref{eq: final kernel estimation}) still contain some noise.
Compared with Figs.~\ref{fig:PSF estimation with priors}(e) and (f),
the new spatial term $\mathcal{C}(k)$ is able to remove noise effectively.

\subsection{Interim Latent Image Estimation}
%\textbf{Interim Latent Image Estimation:}
In this deconvolution stage, we focus on the sharp edges restoration from the
blurred image. Thus, we employ the anisotropic TV model to guide the latent image restoration.
It can be written as
\begin{eqnarray}\label{eq:image deblurring-1}
\min_I \|B-k*I\|_2^2+\lambda_c \|\nabla I\|_1.
\end{eqnarray}
%%%%%%%%%%%%%%%%%%%%%%%%%%%%%%%%%%%%%%%%%%%%%%%%%%%%%%%%%%%%%%%%%%%%%%%%%%%%%%%%%%%%%%%
%%%%%%%%%%%%%%%%%%%%%%%%%%%%%%%%%%%%%%%%%%%%%%%%%%%%%%%%%%%%%%%%%%%%%%%%%%%%%%%%%%%%%%%
We use the IRLS method to solve model~(\ref{eq:image deblurring-1}).
During the iterations we empirically run IRLS for 3 iterations, with
the weights being computed from the recovered image of the previous iterations.
In the inner IRLS system, we use 30 conjugate gradient (CG) iterations.
%%%%%%%%%%%%%%%%%%%%%%%%%%%%%%%%%%%%%%%%%%%%%%%%%%%%%%%%%%%%%%%%%%%%%%%%%%%%%%%%%%%%%%%
%%%%%%%%%%%%%%%%%%%%%%%%%%%%%%%%%%%%%%%%%%%%%%%%%%%%%%%%%%%%%%%%%%%%%%%%%%%%%%%%%%%%%%%

\subsection{Multi-scale Implementation Strategy}
%\textbf{Multi-scale Implementation Strategy:}
To get a better reasonable solution and deal with large blur kernels,
we adopt multi-scale estimation of the kernel using a coarse-to-fine pyramid of image
resolutions which is similar to that in~\cite{Cho/et/al}.
In building the pyramid, we use a downsampling factor of $\frac{\sqrt{2}}{2}$.
The number of pyramid levels is adaptively determined by the size of blur kernel
so that the blur kernel at the coarsest level has a width or height of around 3 to 7 pixels.
%We use a ratio of $\sqrt{2}$ in each adjacent pyramid levels.
At each pyramid level, we perform a few iterations.

%%%%%%%%%%%%%%%%%%%%%%%%%%%%%%%%%%%%%%%%%%%%%%%%%%%%%%%%%%%%%%%%%%%%%%%%%%%%%%%%%%%%%%%
Based on above analysis, our whole kernel estimation algorithm is summarized in Alg.~\ref{alg:jkernel estimation process alg}.
%\vspace*{8pt}
%%%%%%%%%%%%%%%%%%%%%%%%%%%%%%%%%%%%%%%%%%%%%%%%%%%%%%%%%%%%%%%%%%%%%%%%%%%%%%%%%%%%%%%%%%%
\begin{algorithm}\caption{Robust Kernel Estimation from Salient Structure Algorithm}
\label{alg:jkernel estimation process alg}
\begin{algorithmic}
\STATE {\textbf{Input:} Blur image $B$ and the size of blur kernel;}
\STATE {Determine the number of image pyramid $n$ according to the size of kernel;}
\FOR {$i = 1 \to n$}
\STATE Downsample $B$ according to the current image pyramid to get $B_i$;
\FOR {$innerItr = 1 \to m$ ($m$ iterations)}
\STATE Select salient edges $\nabla S$ according to Eq. (\ref{eq:shock structure model});
\STATE Estimate kernel $k$ according to Alg.~\ref{alg:2};
\STATE Estimate latent image $I_i$ according to model~(\ref{eq:image deblurring-1});
\STATE $t \gets t/1.1$, $\theta \gets \theta/1.1$;
%\STATE Check condition: $\|k$;
\ENDFOR
\STATE Upsample image $I_i$ and set $I_{i+1} \gets I_i$;
\ENDFOR
\STATE \textbf{Output:} Blur kernel $k$.
\end{algorithmic}
\end{algorithm}
%%%%%%%%%%%%%%%%%%%%%%%%%%%%%%%%%%%%%%%%%%%%%%%%%%%%%%%%%%%%%%%%%%%%%%%%%%%%%%%%%%%%%%%%%%%
%\vspace*{-8pt}

%-------------------------------------------------------------------------
%%%%%%%%%%%%%%%%%%%%%%%%%%%%%%%%%%%%%%%%%%%%%%%%%%%%%%%%%%%%%%%%%%%%%%%%%%%%%%%%%%%%%%%
\subsection{Analysis on Kernel Estimation}
\label{sec: More Analysis on Kernel Estimation}
In this subsection we provide more insightful analysis on the structure
selection method and the robust kernel estimation model.
\subsubsection{Effectiveness of the Proposed Structure Selection Method}
\label{secsec: Effectiveness of Our Structure Selection Method}
Inaccurate sharp edges
will induce noisy or even wrong kernel estimates which further deteriorate the final recovered images.
In this subsection, we demonstrate the effectiveness of salient edges $\nabla S$
%and provide some analysis about kernel estimation method
via some examples.

\begin{figure}
\setlength{\abovecaptionskip}{-0.3cm} %缩小caption和图像之间的距离
\setlength{\belowcaptionskip}{-0.4cm} %缩小caption和下方文字的距离
\begin{center}
\includegraphics[angle=0, width=0.95\textwidth]{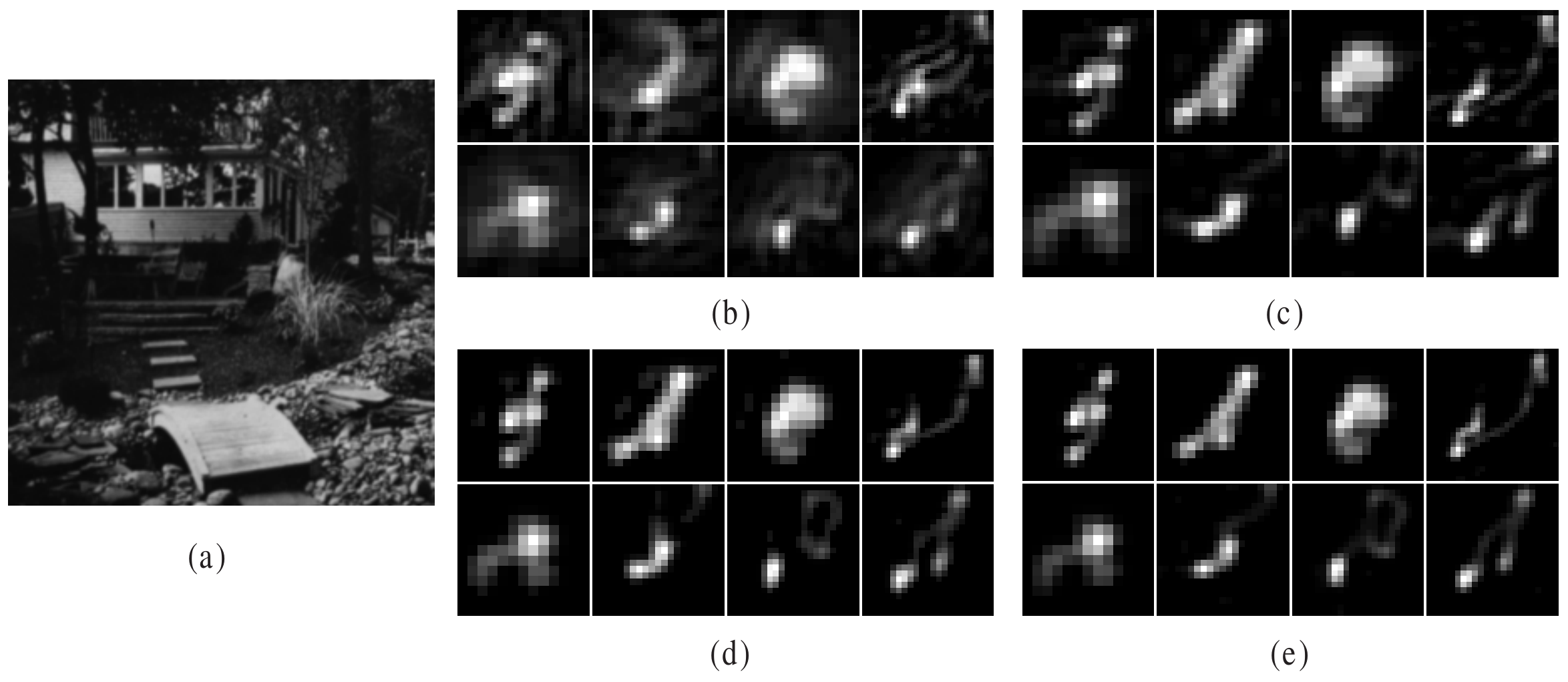}
\end{center}
\caption{Comparison of results with and without model~(\ref{eq:improved TV model}).
(a) The ground truth image.
(b) Kernel estimates without adopting model~(\ref{eq:improved TV model}).
(c) Kernel estimates of~\cite{Cho/et/al}.
(d) Kernel estimates of~\cite{Xu/et/al}.
(e) Kernel estimates with model~(\ref{eq:improved TV model}).
}
\label{fig:PSF estimation}
\end{figure}

As mentioned in Section~\ref{sec:introduction} and Section~\ref{subsec: comppute structure},
image details will damage the kernel estimation. Therefore,
we use salient edges $\nabla S$ to estimate kernels.
%The Salient Edges $\Nabla S$ Are Extracted From Interim Latent Images, And It Contains
%The Gradient With Large Pixel Values.
%Our kernel estimation mainly relies on the reliable salient edges $\nabla S$.
To verify the validity of $\nabla S$,
we perform several experiments by using the data from~\cite{Levin/CVPR2009}.
Furthermore, to emphasize the fact that tiny structures damage the kernel estimation,
we select the dataset with rich details from~\cite{Levin/CVPR2009}
(shown in Fig.~\ref{fig:PSF estimation}(a)).

Fig.~\ref{fig:PSF estimation} shows an example that demonstrates the
effectiveness of model~(\ref{eq:improved TV model}) in the kernel estimation process.
Due to the proposed structure selection mechanism, the
kernel estimates (shown in Fig.~\ref{fig:PSF estimation} (e))
outperform those without adopting model~(\ref{eq:improved TV model}) (i.e., Fig.~\ref{fig:PSF estimation} (b)).
Compared with other structure selection methods~\cite{Cho/et/al,Xu/et/al},
our method outperforms better.

\begin{figure}
\begin{center}
\includegraphics[angle=0, width=0.8\textwidth]{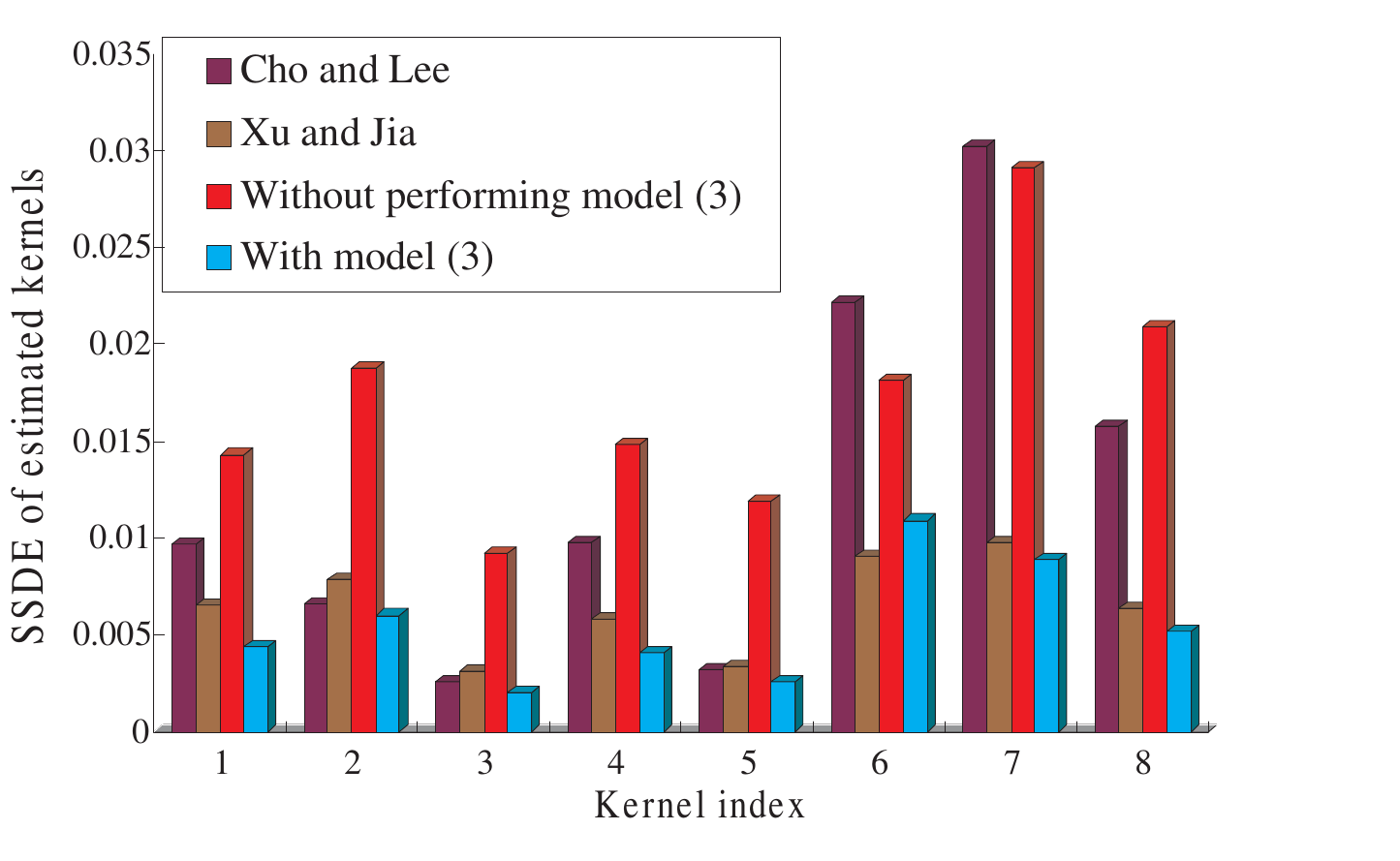}
\end{center}
\vspace{-8mm}
\caption{Comparison of kernel estimation results in terms of SSDE.}
\label{fig:PSF estimation SSDE}
\end{figure}

In Fig.~\ref{fig:PSF estimation SSDE}, Sum of Squared Differences Error (SSDE)
is employed to compare the estimation accuracy for the blur kernels in Fig.~\ref{fig:PSF estimation}.
One can see that the accuracy of kernel estimation by the proposed method has been
greatly improved.

\begin{figure}
\begin{center}
\includegraphics[angle=0, width=0.7\textwidth]{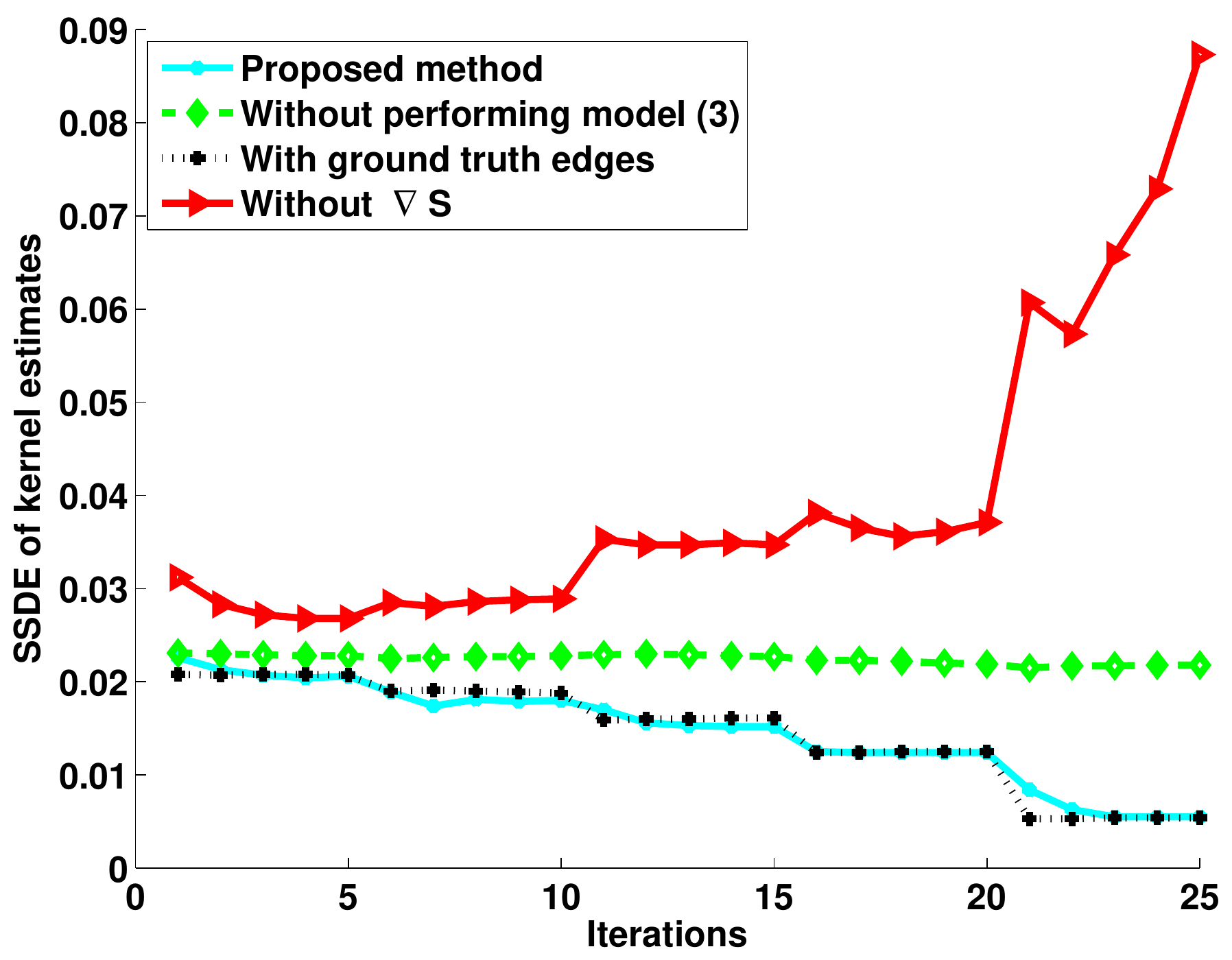}
\end{center}
\caption{Importance of salient edges $\nabla S$ and the effectiveness of our whole kernel estimation algorithm.
The red curve shows the kernel estimation errors without adopting any structure selection strategies.
The dashed green curve shows the kernel estimation errors without adopting model~(\ref{eq:improved TV model})
in the proposed structure selection strategy.
The cyan curve shows the kernel estimation errors with the proposed structure selection strategy.
The dotted black curve shows the kernel estimation errors with the proposed structure selection strategy,
while the salient edges are extracted from the ground truth image.
}
\label{fig:PSF estimation -curves}
\end{figure}

To further demonstrate the importance of salient edges $\nabla S$ and the effectiveness of
our whole kernel estimation algorithm,
we choose an example (i.e., the ``\textmd{im02\_ker08}" test case in the dataset in~\cite{Levin/CVPR2009})
to conduct another experiment shown in Fig.~\ref{fig:PSF estimation -curves}.
%As can be seen from Fig.~\ref{fig:PSF estimation -curves},
%the SSDE values of kernel estimates without adopting salient edges $\nabla S$
As we do not use salient edges $\nabla S$,
the SSDE values of kernel estimates
(i.e., the red curve in Fig.~\ref{fig:PSF estimation -curves})
are increasing with the iterations.
In contrast, the results with $\nabla S$
(i.e., the cyan curve and the dashed green curve in Fig.~\ref{fig:PSF estimation -curves}) are better.
This further demonstrates the importance of salient edges.
Compared the cyan curve with the dashed green curve, the quality of kernel estimates
that are generated by model~(\ref{eq:improved TV model}) has been greatly improved.
In addition, its accuracy and convergency are better than those without adopting
model~(\ref{eq:improved TV model}).
This is also in line with our understanding (Illustrated in Section~\ref{sec:introduction}
and Section~\ref{subsec: comppute structure}).
From the cyan curve and the dotted black curve, one can see that our salient edges $\nabla S$
perform comparably to the salient edges that are extracted from the ground truth images.
This further verifies the validity of our structure selection method.
\subsubsection{Effectiveness of the Proposed Kernel Estimation Model}
\label{secsec: Effectiveness of Our Kernel Estimation Model}
Although salient edges are very important, a robust kernel estimation model also plays a
critical part in the kernel estimation process.
Thus, we propose model~(\ref{eq: improved final kernel estimation}) to estimate kernels.
%An effective constraint Eq.~(\ref{eq: constraint}) is introduced to remove
%The robust model~(\ref{eq: improved final kernel estimation}) is able to remove
%noise that is caused by some unreliable salient edges.
%noise when reliability of some salient edges are not guaranteed.
The results shown in Fig.~\ref{fig:PSF estimation} illustrate its effectiveness to some extent.
The kernel estimates by~\cite{Cho/et/al,Xu/et/al} contain some obvious noise
and the continuity of some kernel estimates has also been destroyed.
However, results shown in Fig.~\ref{fig:PSF estimation} (e) demonstrate
that model~(\ref{eq: improved final kernel estimation}) not only can remove noise
but also preserve the continuity of the blur kernels.
%Results shown in Fig.~\ref{fig:PSF estimation with priors} demonstrate such a possibility.

To provide a more insightful illustration,
we use the same dataset shown in Fig.~\ref{fig:PSF estimation}(a) to demonstrate
the effectiveness of our kernel estimation model.

\begin{figure}
\begin{center}
\includegraphics[angle=0, width=0.5\textwidth]{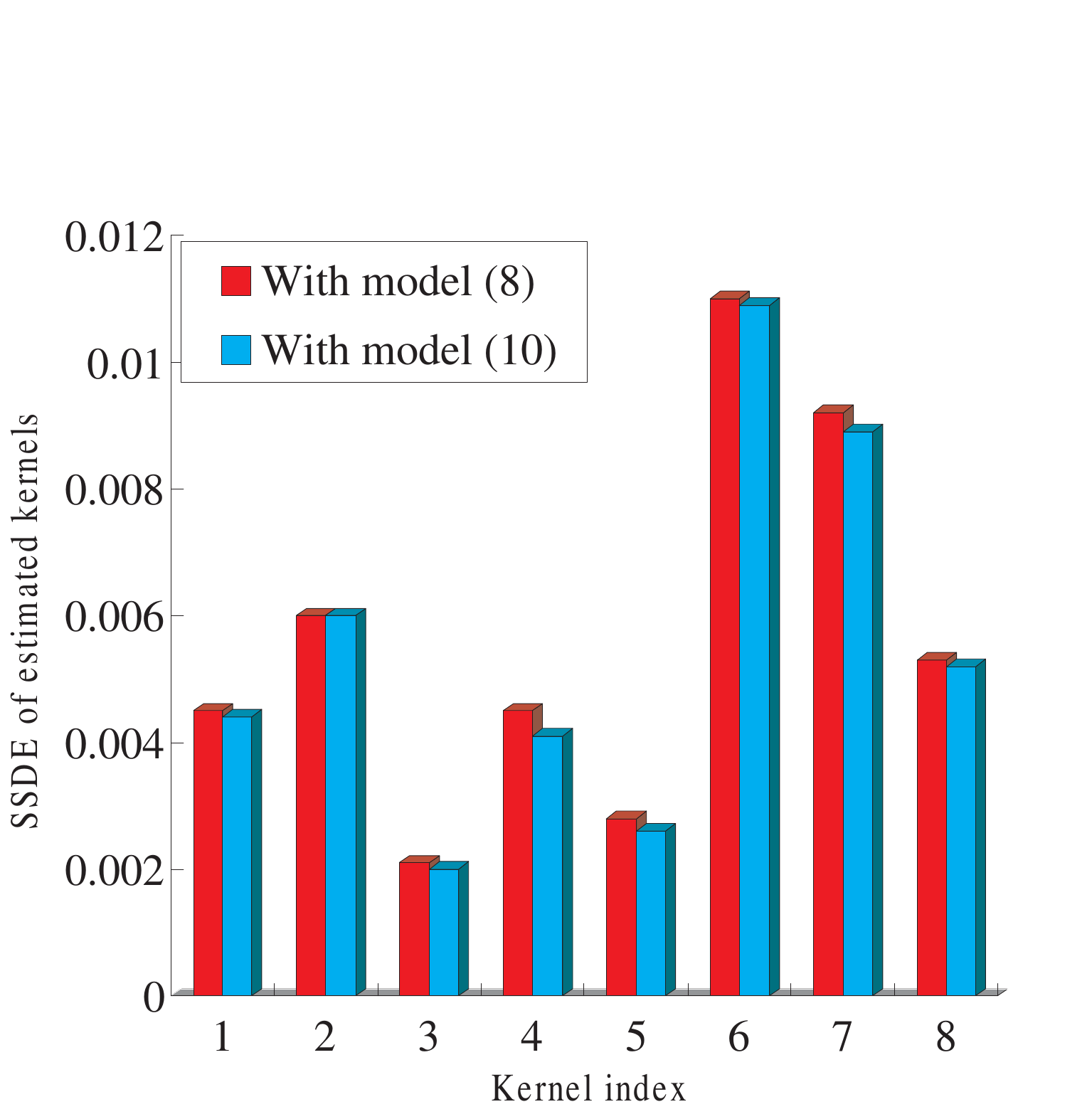}
\end{center}
\vspace{-10mm}
\caption{The effectiveness of our kernel estimation model.
}
\label{fig:kernel-smooth}
\end{figure}

Fig.~\ref{fig:kernel-smooth} shows the comparison of kernel estimation results in terms of SSDE.
Due to the influence of the constraint Eq.~(\ref{eq: constraint}), the accuracy of estimated kernels
is much higher.
More illustrative examples are included in supplemental material.

\section{Final Latent Image Estimation}

Model~(\ref{eq:image deblurring-1}) may lead to the stair-casing effect and destroy textures.
To overcome this problem, some adaptive regularization terms have been proposed and proved
to be effective in edge-preserving~\cite{edge/preserving/tog/08}.
Inspired by this idea, we utilize our predicted structure to
guide the latent image restoration.
%The predicted structure $\nabla S$ is utilized as the weight of spatial prior.
%Thus,
Our final model for latent image restoration is defined as
 \begin{eqnarray}\label{eq:final image deblurring}
\min_I \|B-k*I\|_2^2+\lambda(\exp(-\|\partial_x S\|^{0.8})\|\partial_x I\|_1 + \exp(-\|\partial_y S\|^{0.8})\|\partial_y I\|_1).
\end{eqnarray}
%where $\beta_i=\exp(-\|S_i(x,y)\|^{0.8})$, $i\in\{x, y\}$.
In model~(\ref{eq:final image deblurring}),
the smoothness requirement is enforced in a spatially varying manner
via the smoothness weights $\exp(-\|\partial_x S\|^{0.8})$ and $\exp(-\|\partial_y S\|^{0.8})$,
which depend on the salient edges $\nabla S$.
Hence, model~(\ref{eq:final image deblurring}) is able to contribute to the edge-preserving.

Model~(\ref{eq:final image deblurring}) can also be solved by the IRLS method efficiently.
We run IRLS for 3 iterations.
At each iterations, the weights for $\partial_x I$ and $\partial_y I$ are defined as
\begin{eqnarray}\label{eq:weight for regularization-1}
\left\{\begin{array}{ll}w_x = \frac{\exp(-|\partial_x S|^{0.8})}{\max \{|\partial_x I|,0.001\}}, \\w_y = \frac{\exp(-|\partial_y S|^{0.8})}{\max \{|\partial_y I|,0.001\}}.\ \end{array}\right.
\end{eqnarray}
We use 100 CG iterations in the inner IRLS system.

Based on above analysis, our deblurring algorithm is summarized in Alg.~\ref{alg:33}.
%\vspace*{8pt}
%%%%%%%%%%%%%%%%%%%%%%%%%%%%%%%%%%%%%%%%%%%%%%%%%%%%%%%%%%%%%%%%%%%%%%%%%%%%%%%%%%%%%%%%%%%
\begin{algorithm}\caption{The Completed Image Deblurring Algorithm}
\label{alg:33}
\begin{algorithmic}
\STATE {\textbf{Input:} Blur image $B$ and the size of blur kernel;}
\STATE {\textbf{Step 1:} Estimate kernel $k$ by Alg.~\ref{alg:jkernel estimation process alg};}
\STATE {\textbf{Step 2:} Estimate the final latent image $I$ according
to model~(\ref{eq:final image deblurring})};
\STATE {\textbf{Output:} Latent image $I$}.
\end{algorithmic}
\end{algorithm}
%%%%%%%%%%%%%%%%%%%%%%%%%%%%%%%%%%%%%%%%%%%%%%%%%%%%%%%%%%%%%%%%%%%%%%%%%%%%%%%%%%%%%%%%%%%
%\vspace*{-8pt}

%-------------------------------------------------------------------------
\section{Experiments}
In this section, we present results of our algorithm and
compare it to the state-of-the-art approaches of~\cite{Fergus/et/al,Shan/et/al,Cho/et/al,Levin/CVPR2011,Xu/et/al,Krishnan/CVPR2011}.
%We compared our method with other state-of-the-art approaches on several challenging examples.
We first introduce some implementation details.
In the kernel estimation, all color images are converted to grayscale ones.
The initialized value of $\theta$ is experimentally set to $1$ based on lots of experiments.
%and $t$ is initialized according to~\cite{Cho/et/al}.
The parameter $\lambda_c$ in model~(\ref{eq:image deblurring-1}) is set to
$0.005$, $\lambda$ in model~(\ref{eq:final image deblurring}) is set to $0.003$, and $\gamma$ in
model~(\ref{eq: improved final kernel estimation}) is set to $0.01$.
%%%%%%%%%%%%%%%%%%%%%%%%%%%%%%%%%%%%%%%%%%%%%%%%%%%%%%%%%%%%%%%%%%%%
In Alg.~\ref{alg:2}, solving model~(\ref{eq: L0 smooth_AM-Xu}) will produce negative values.
%We adopt Yuan {\it et al.}'s method~\cite{tog/YuanSQS07} to remove negative values of blur kernels.
We project the estimated blur kernel onto the constraints (i.e., setting negative elements to 0 and renormalizing).
In Alg.~\ref{alg:jkernel estimation process alg}, we empirically set the inner iteration $m=5$.
%%%%%%%%%%%%%%%%%%%%%%%%%%%%%%%%%%%%%%%%%%%%%%%%%%%%%%%%%%%%%%%%%%%%
In the final deconvolution process, each color channel
is separately processed. 
%The MATLAB codes of Alg.~\ref{alg:jkernel estimation process alg} are available at~\url{https://www.dropbox.com/s/eixi8a2nsg15mhk/Deblurring_code_v2.zip}.
\subsection{Experimental Results and Evaluation}
\begin{figure}
\centering
\includegraphics[angle=0, width=0.98\textwidth]{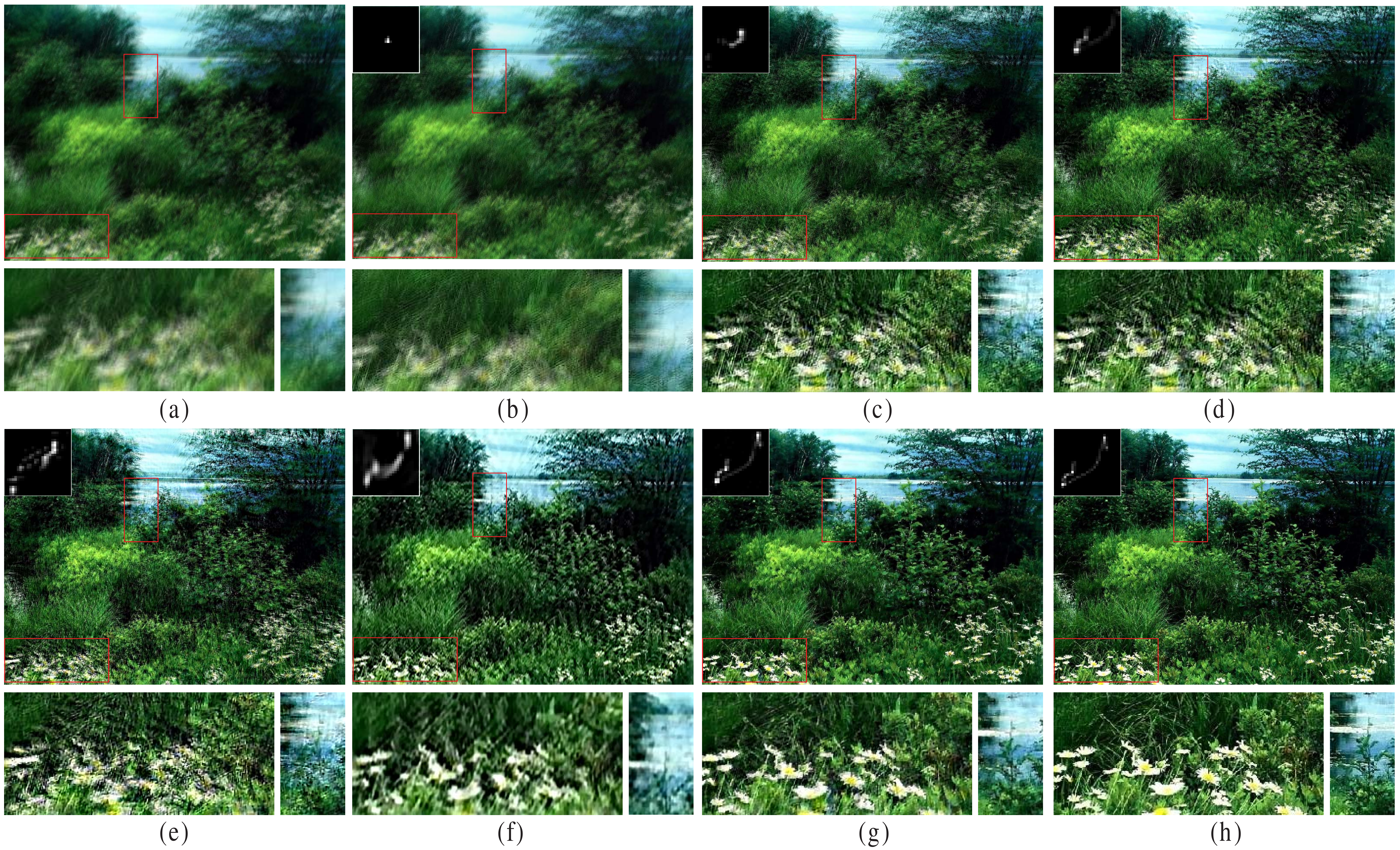}
\caption{Small tiny details such as the grass and leaves are contained in the image.
(a) Blurred image.
(b) Result of Fergus {\it et al.}~\cite{Fergus/et/al}.
(c) Result of Shan {\it et al.}~\cite{Shan/et/al}.
(d) Result of Cho and Lee~\cite{Cho/et/al}.
(e) Result of Xu and Jia~\cite{Xu/et/al}.
(f) Result of Krishnan {\it et al.}~\cite{Krishnan/CVPR2011}.
(g) Our result.
(h) The ground truth result.
The size of motion blur kernel is $27\times 27$.}
\label{fig:Compare grass result}
\end{figure}

We first use a synthetic example shown in Fig.~\ref{fig:Compare grass result}(a)
to prove the effectiveness of our method. The blurred image
contains rich textures and small details, such as flowers,
leaves, and grass which increase the difficulty in kernel
estimation.
%We compare our method with state-of-the-art motion deblurring methods
%~\cite{Fergus/et/al,Shan/et/al,Cho/et/al,Xu/et/al,Krishnan/CVPR2011}.
Methods of Fergus {\it et al.}~\cite{Fergus/et/al} and Shan {\it et al.}~\cite{Shan/et/al}
fail to provide correct kernel estimation results, and their deblurred results still contain some
blur and ringing artifacts.
Other methods~\cite{Xu/et/al,Cho/et/al,Krishnan/CVPR2011} provide deblurred results with some ringing artifacts
due to the imperfect kernel estimation results.
However, our results shown in Fig.~\ref{fig:Compare grass result}(g) perform well both
in the kernel estimation and the final latent image restoration.

In Table~\ref{tab:error table}, we employ SSDE and PSNR (Peak Signal to Noise Ratio) to compare the estimation accuracy for
the blur kernels and the restored images in Fig.~\ref{fig:Compare grass result}, respectively.
Our method provides higher PSNR value for the restored image and lower SSDE value for the kernel estimate.
%----------------------------------------------------------------------------------------------------
\setlength{\tabcolsep}{4pt}
\begin{table}[h]
\centering
\caption{\label{tab:error table} Comparison of estimated results in Fig.~\ref{fig:Compare grass result}.}
 \begin{tabular}{lclllllllllc}
  \toprule
  \qquad Methods &\cite{Fergus/et/al} & \cite{Shan/et/al} & \cite{Cho/et/al} & \cite{Xu/et/al} & \cite{Krishnan/CVPR2011} & Ours\\
  \midrule
  PSNR of images & 15.45 & 14.78 & 14.44 & 13.47 & 15.33 & \textbf{19.92}\\
  SSDE of kernels & 0.2654 & 0.0329 & 0.0298 & 0.0301 & 0.0292 & \textbf{0.0021}\\
  \bottomrule
 \end{tabular}
\end{table}
\setlength{\tabcolsep}{1.4pt}

\begin{figure}
\centering
\includegraphics[angle=0, width=0.98\textwidth]{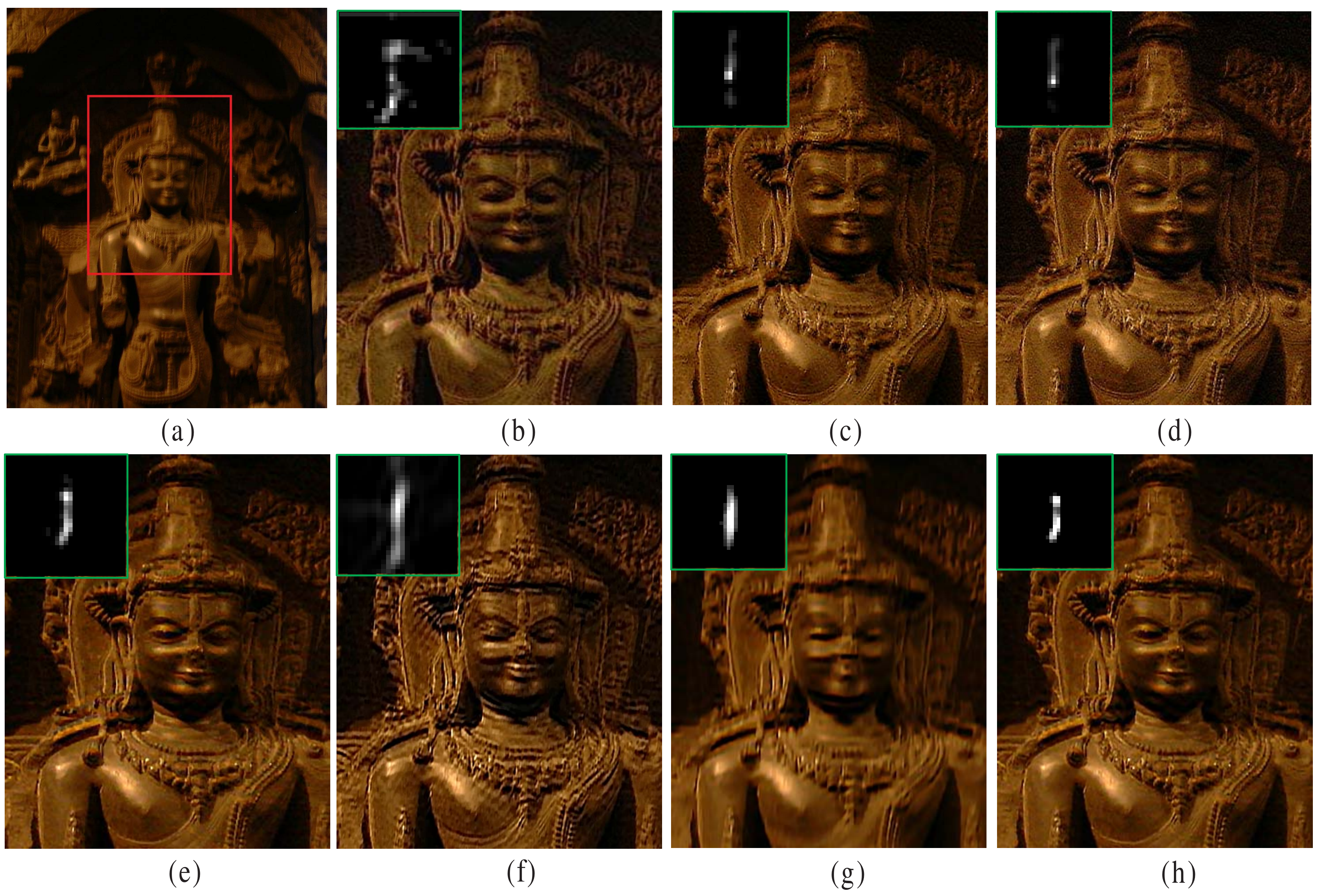}
\caption{Comparison of results with and without using image structures.
(a) Blurred image.
(b) - (h) are deblurred results cropped from the red box in (a).
(b) Result of Fergus {\it et al.}~\cite{Fergus/et/al}.
(c) Result of Shan {\it et al.}~\cite{Shan/et/al}.
(d) Result of Cho and Lee~\cite{Cho/et/al}.
(e) Result of Xu and Jia~\cite{Xu/et/al}.
(f) Result of Krishnan {\it et al.}~\cite{Krishnan/CVPR2011}.
(g) Result without performing model~(\ref{eq:improved TV model}).
(h) Result with model~(\ref{eq:improved TV model}).
The result (h) of our method is the best.
}
\label{fig:Compare fo result}
\end{figure}

We then test the effectiveness of our structure selection model~(\ref{eq:improved TV model}).
Fig.~\ref{fig:Compare fo result}(a) is a real captured image presented in~\cite{Fergus/et/al}.
The deblurred results of~\cite{Fergus/et/al,Shan/et/al,Cho/et/al} contain some noise.
The results of Xu and Jia~\cite{Xu/et/al} perform better, but the kernel estimate still
contains some noise. Our method shown in Fig.~\ref{fig:Compare fo result}(h) performs better
in kernel estimation and latent image restoration.
Fig.~\ref{fig:Compare fo result}(g) shows our result without performing model~(\ref{eq:improved TV model}).
Compared to the result shown in Fig.~\ref{fig:Compare fo result}(h),
its quality is lower, indicating the importance of structures in estimating kernels.

\begin{figure}[!t]
\centering
\includegraphics[angle=0, width=0.98\textwidth]{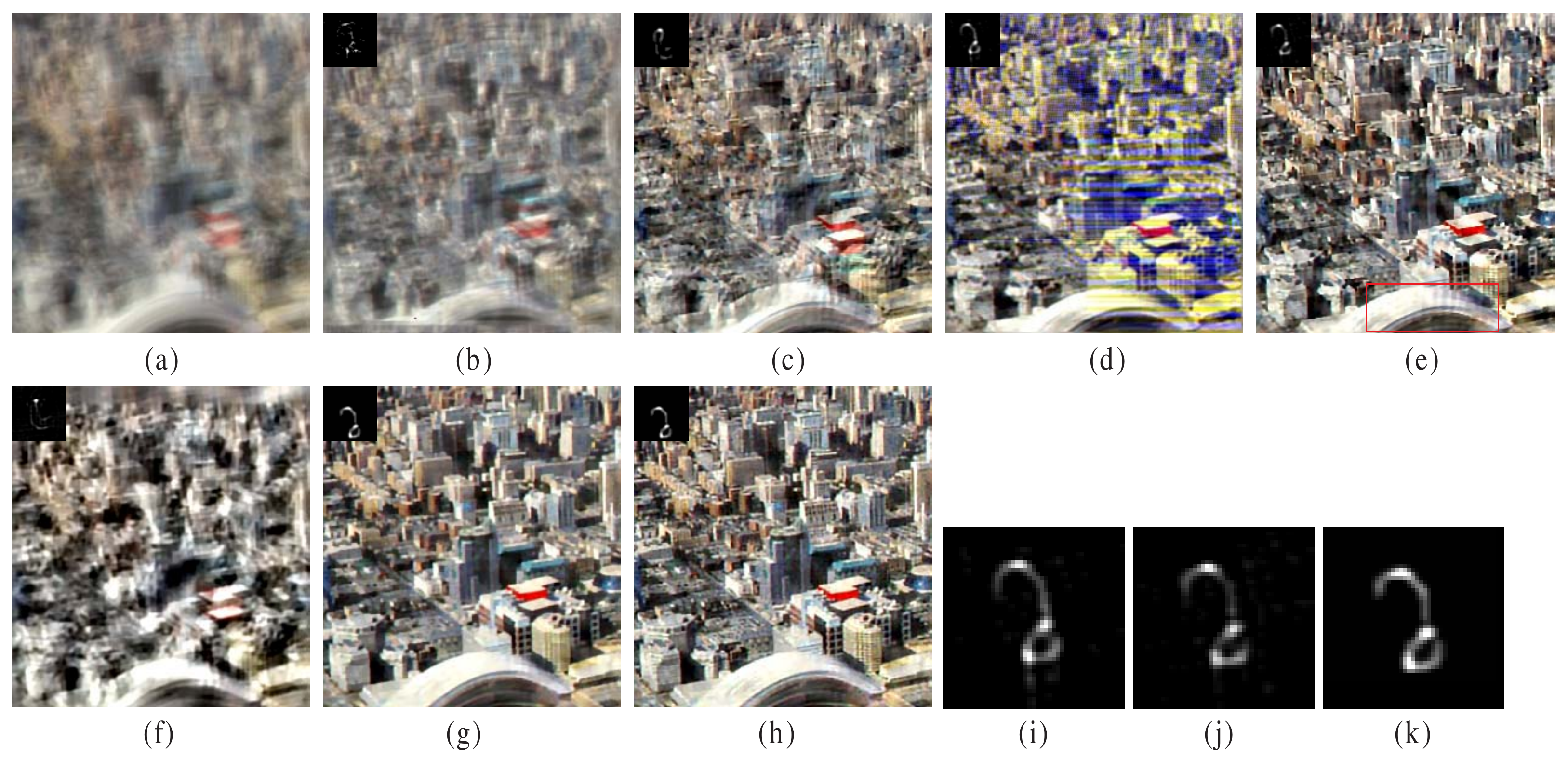}
\caption{A challenging example with much tiny structures.
These tiny structures greatly increase the difficulty of kernel estimation.
(a) Blurred image.
(b) Result of Fergus {\it et al.}~\cite{Fergus/et/al}.
(c) Result of Shan {\it et al.}~\cite{Shan/et/al}.
(d) Result of Cho and Lee~\cite{Cho/et/al}.
(e) Result of Xu and Jia~\cite{Xu/et/al}.
(f) Result of Krishnan {\it et al.}~\cite{Krishnan/CVPR2011}.
(g) - (h): Our results.
Deblurred results in (g) and (h) are generated by model~(\ref{eq:image deblurring-1}) and~(\ref{eq:final image deblurring}), respectively.
(i) Kernel estimate in~\cite{Cho/et/al}.
(j) Kernel estimate in~\cite{Xu/et/al}.
(k) Our kernel estimation result.
The size of blur kernel is $45\times 45$.
}
\label{fig:challenge example with much tinny structure}
\end{figure}

For real images with rich textures and small details, our method
can still achieve good results. Fig.~\ref{fig:challenge example with much tinny structure}(a)
shows a challenging example with tiny
structures in the blurred image (published in~\cite{Xu/et/al}). The
methods of Fergus {\it et al.}~\cite{Fergus/et/al}, Shan
{\it et al.}~\cite{Shan/et/al}, and Krishnan
{\it et al.}~\cite{Krishnan/CVPR2011} fail to provide the correct deblurred
results and the kernel estimation results.
The method of Cho and Lee~\cite{Cho/et/al} is able to estimate the blur kernel,
but the deblurred contains some extra artifacts.
The estimated kernel of Xu and Jia~\cite{Xu/et/al} is better, but the final deblurred result still
contains some visual artifacts (shown in the red box in Fig.~\ref{fig:challenge example with much tinny structure}(e)).
Moreover, the kernel estimation result contains some obvious noise (Fig.~\ref{fig:challenge example with much tinny structure}(j)).
Our method outperforms these methods both in the kernel estimation and the latent image
restoration. Compared with Fig.~\ref{fig:challenge example with much tinny structure}(g) and
Fig.~\ref{fig:challenge example with much tinny structure}(h), our simple adaptive weighted
spatial prior can preserve more sharp edges and finer textures in the latent image.

\begin{figure}[!t]
\centering
\includegraphics[angle=0, width=0.98\textwidth]{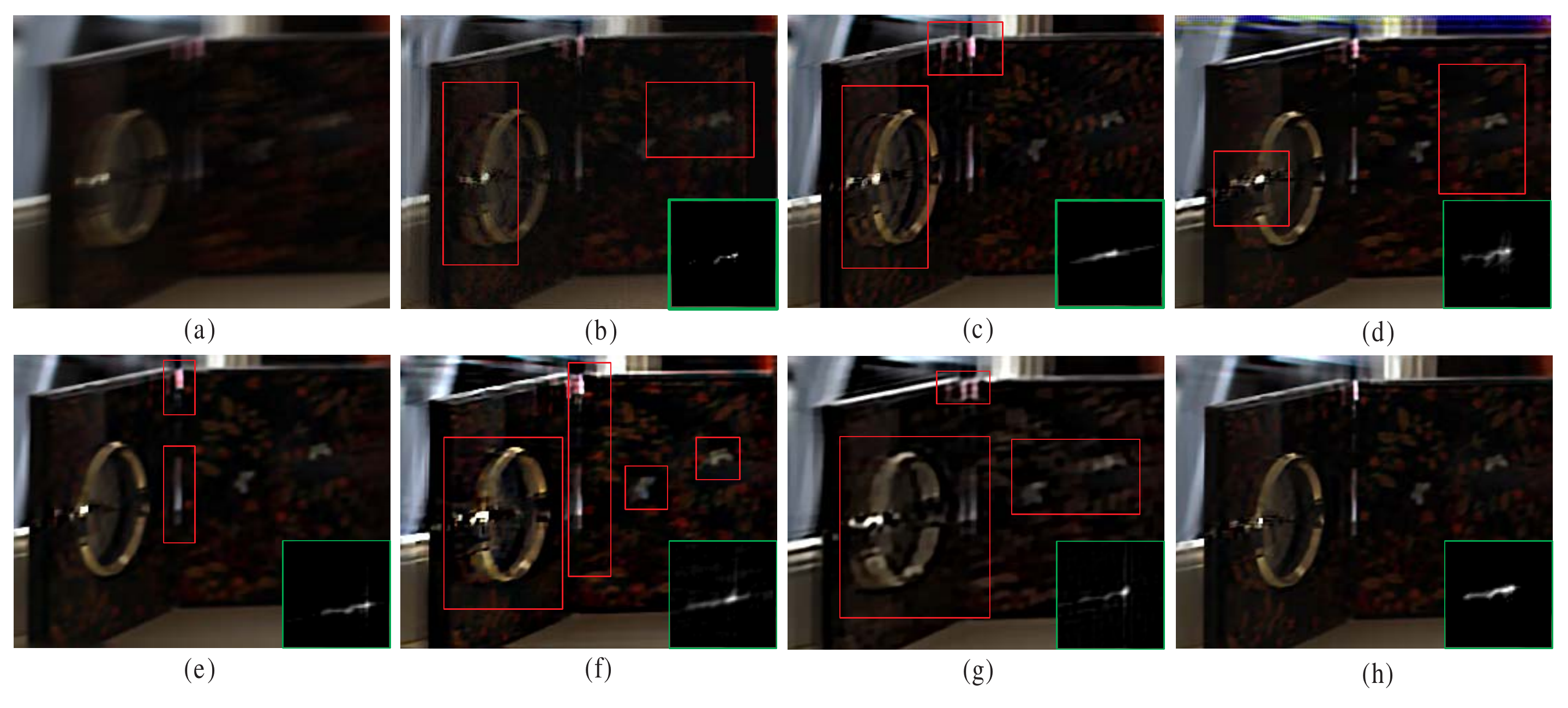}
\caption{A large blur kernel estimation example.
(a) Blurred image.
(b) Result of Fergus {\it et al.}~\cite{Fergus/et/al}.
(c) Result of Shan {\it et al.}~\cite{Shan/et/al}.
(d) Result of Cho and Lee~\cite{Cho/et/al}.
(e) Result of Levin {\it et al.}~\cite{Levin/CVPR2011}.
(f) Result of Xu and Jia~\cite{Xu/et/al}.
(g) Result of Krishnan {\it et al.}~\cite{Krishnan/CVPR2011}.
(h) Our result.
The red boxes shown in (b) - (g) still contain some ringing artifacts or blur.
Our estimated blur kernel size is $53\times 53$.}
\label{fig:Compare our result}
\end{figure}

Another important advantage of our method is that it can
deal with large blur kernels. The photo
in Fig.~\ref{fig:Compare our result}(a) is captured by ourselves,
whose motion blur is quite large. The method of~\cite{Cho/et/al}
performs better than that of~\cite{Xu/et/al}, but the kernel
estimation result still contains some noise, and the deblurred result is
inaccurate in the red box. Due to the large blur, methods of~\cite{Fergus/et/al,Shan/et/al,Krishnan/CVPR2011}
cannot produce correct kernel estimation results either and their deblurred results still
contain some obvious blur and ringing artifacts (e.g., the part in the red boxes).
Levin {\it et al.}~\cite{Levin/CVPR2011}'s method provides better estimated results, but the estimated kernel
still contains some noise and the deblurred result also contains some
blur (e.g., the parts in the red boxes in Fig.~\ref{fig:Compare our result}(e))
Our approach, however, generates a better kernel estimate and the deblurred result is also
visually comparable.

\begin{figure}[!t]
\centering
\includegraphics[angle=0, width=0.98\textwidth]{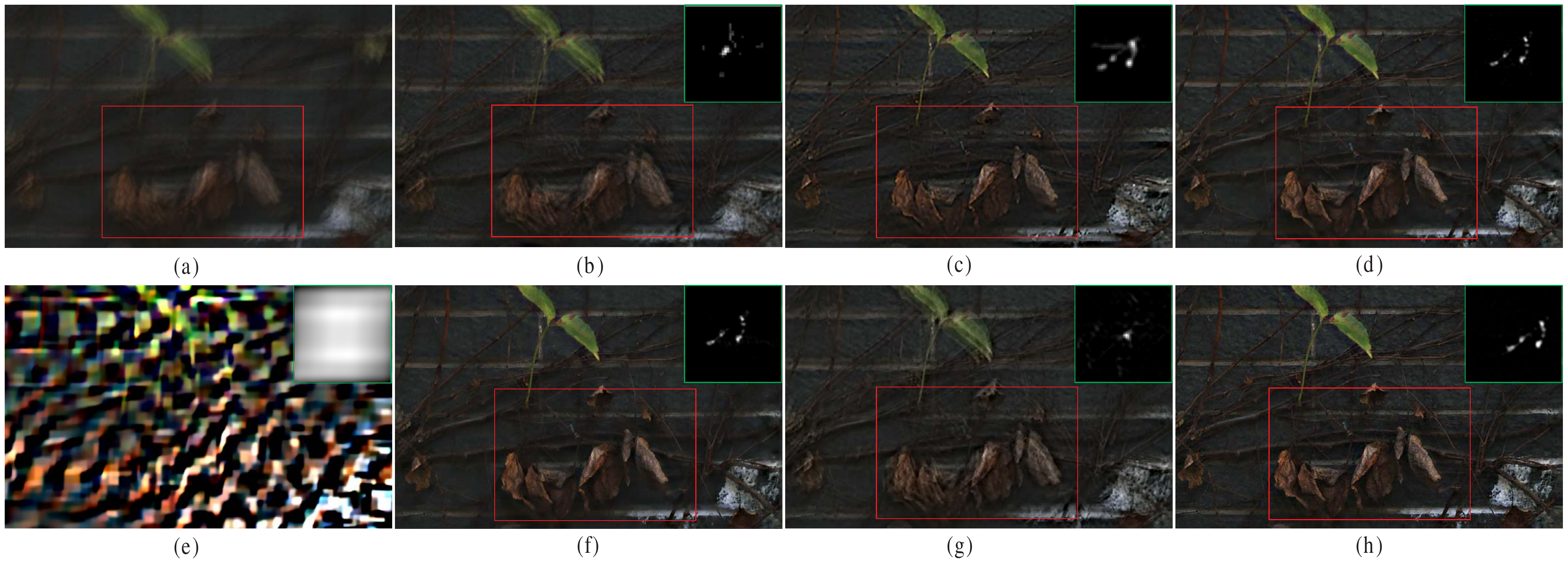}
\caption{Another large blur kernel estimation example.
(a) Blurred image.
(b) Result of Fergus {\it et al.}~\cite{Fergus/et/al}.
(c) Result of Shan {\it et al.}~\cite{Shan/et/al}.
(d) Result of Cho and Lee~\cite{Cho/et/al}.
(e) Result of Levin {\it et al.}~\cite{Levin/CVPR2011}.
(f) Result of Xu and Jia~\cite{Xu/et/al}.
(g) Result of Krishnan {\it et al.}~\cite{Krishnan/CVPR2011}.
(h) Our result.
Our estimated blur kernel size is $99\times 99$.}
\label{fig:Compare of leaf result}
\end{figure}

Fig.~\ref{fig:Compare of leaf result} shows another example with
large motion blur. The blurred image (Fig.~\ref{fig:Compare of leaf result}(a)) also contains small
details. Due to the large blur, methods~\cite{Fergus/et/al,Krishnan/CVPR2011,Levin/CVPR2011}
cannot provide reasonable results. The deblurred result of~\cite{Shan/et/al} still
contains some noise and ringing artifacts.
Results of~\cite{Cho/et/al,Xu/et/al} still contain some blur.
Our method, however, provides a clearer image with finer textures.

\textbf{Evaluation on the Synthetic Dataset~\cite{Levin/CVPR2009}:}
We perform quantitative evaluation of our kernel estimation method
by using the dataset from Levin {\it et al.}~\cite{Levin/CVPR2009},
and compare our kernel estimation results with the state-of-the-art
blind deblurring algorithms by Fergus {\it et al.}~\cite{Fergus/et/al},
Shan {\it et al.}~\cite{Shan/et/al},
Cho and Lee~\cite{Cho/et/al}, Xu and Jia~\cite{Xu/et/al}, Krishnan {\it et al.}~\cite{Krishnan/CVPR2011},
and Levin {\it et al.}'s latest method~\cite{Levin/CVPR2011}.
For evaluation with each test case, we follow the method used
in~\cite{Levin/CVPR2009}. The kernel estimation results
of~\cite{Fergus/et/al,Shan/et/al,Cho/et/al,Xu/et/al,Krishnan/CVPR2011,Levin/CVPR2011} are all generated by
using the authors' source codes or executable programs downloaded online.
Then, the deblurred results are obtained by using Levin {\it et al.}'s~\cite{Levin/CVPR2011} matlab
function \verb"deconvSps.m" with the same parameter settings.
The error metric is also the same as~\cite{Levin/CVPR2009} and it is defined as
\begin{eqnarray}\label{eq:error-metric}
\mathcal{E} = \frac{\|I_r - I_g\|_2^2}{\|I_t - I_g\|_2^2},
\end{eqnarray}
where $I_r$ and $I_t$ are the restored images with the estimated kernel and ground truth kernel, respectively,
and $I_g$ is the ground truth image.

\begin{figure}
\begin{center}
\includegraphics[angle=0, width=0.7\textwidth]{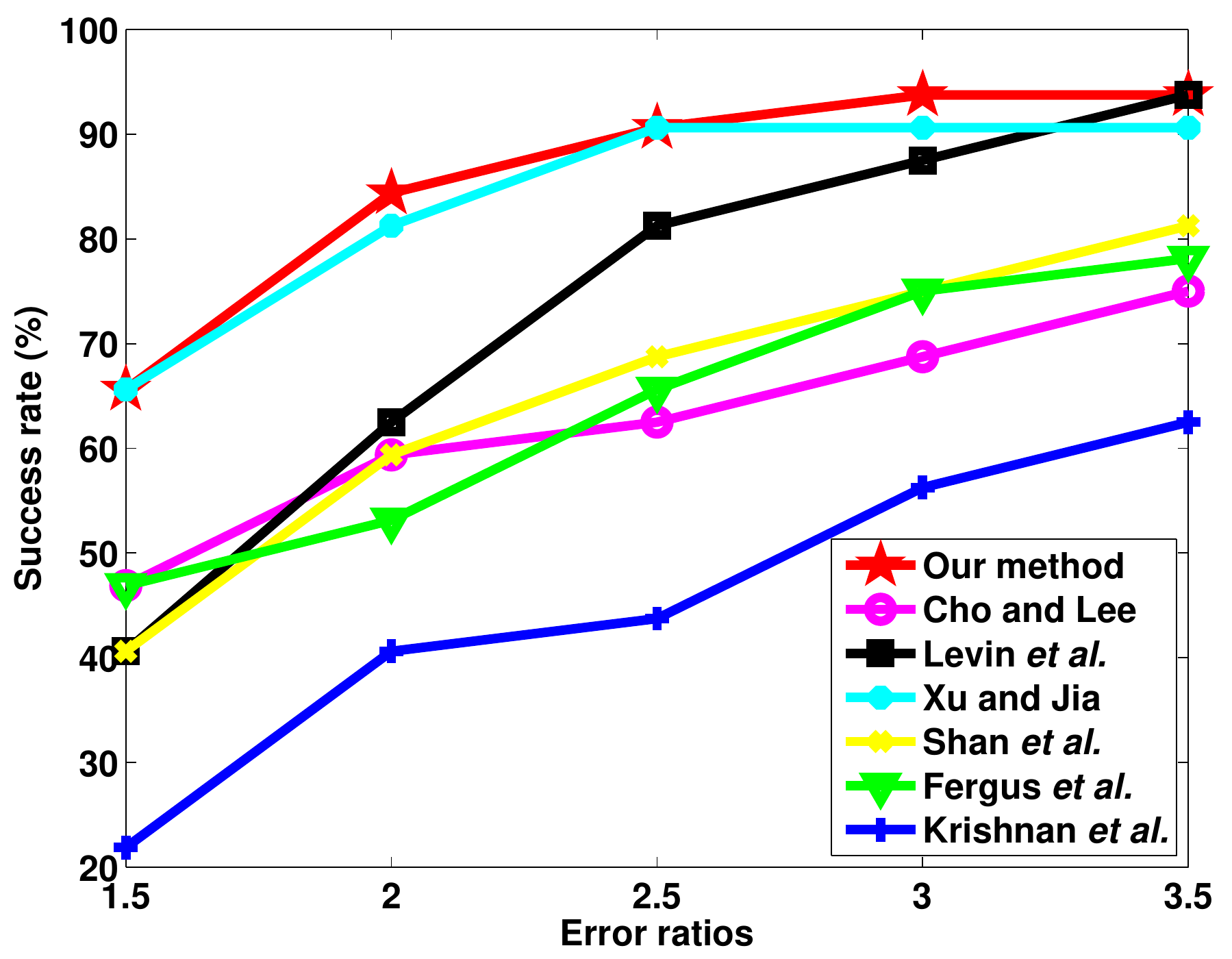}
\end{center}
\caption{Cumulative histogram of the deconvolution error ratio across test examples.}
\label{fig:error-ratios}
\end{figure}

In Fig.~\ref{fig:error-ratios}, we plot the cumulative histograms of deconvolution
error ratios in the same way as~\cite{Levin/CVPR2011}.
In the x-axis, a number of $n$ shows the percentage
of test cases whose deconvolution error ratios are below $n$.
Our method provides more reliable results than others.
More comparison results can be found in our supplementary materials.
%%%%%%%%%%%%%%%%%%%%%%%%%%%%%%%%%%%%%%%%%%%%%%%%%%%%%%%%%%%%%%%%%%%%%%%%%%%%%
%%%%%%%%%%%%%%%%%%%%%%%%%%%%%%%%%%%%%%%%%%%%%%%%%%%%%%%%%%%%%%%%%%%%%%%%%%%%%

\subsection{Computational Cost}
In the kernel estimation process, we should iteratively solve models~(\ref{eq: improved final kernel estimation})
and~(\ref{eq:image deblurring-1}) which involve a few matrix-vector or convolution operations.
For the computational time,
our Matlab implementation spends about 2 minutes to estimate a $27 \times 27$ kernel
from a $255 \times 255$ image with an Intel Xeon CPU@2.53GHz and 12GB RAM,
while methods~\cite{Fergus/et/al,Krishnan/CVPR2011,Levin/CVPR2011} need about 7 minutes, 3 minutes, and
4 minutes, respectively\footnote{The computational time is computed by using the author's matlab source code.}.
The algorithm~\cite{Shan/et/al} implemented in C++ spends about 50 seconds.
Compared with~\cite{Cho/et/al,Xu/et/al}, our method needs more computational time
due to involving non-convex models in kernel estimation.
However, in our kernel estimation process,
both the kernel estimation step and latent image restoration step involve the CG method.
Thus, we believe that our method is amenable to speedup with GPU acceleration by the strategy in~\cite{Cho/et/al}.

\subsection{Handling Blurred Images with Outliers}
Outliers in the blurred image will increase the difficulty in kernel estimation and latent image restoration.
Recent works~\cite{huiji/tip/JiW12,Cho/iccv11,Whyte/iccv2011/workshop} proposed robust non-blind
deblurring methods to deal with outliers.
When dealing with real blurred images with outliers,
they used kernel estimation methods, e.g.,~\cite{Cho/et/al,Xu/et/al}, to estimate blur kernels
and then applied their methods to obtain a better deblurred result.

According to the strategies described in~\cite{huiji/tip/JiW12,Cho/iccv11,Whyte/iccv2011/workshop},
our kernel estimation method can be applied to the images with outliers which distribute non-uniformly in the
blurred image.
Specifically, we use our kernel estimation method to estimate a blur kernel from an image patch without obvious outliers
and then adopt the non-blind deblurring method~\cite{Cho/iccv11} to restore the latent image.
To demonstrate the effectiveness of our kernel estimation method,
we choose an example from~\cite{Tai/noise/cvpr12} and compare our method with~\cite{Tai/noise/cvpr12},
which is specialized on dealing with noise.
\begin{figure}[h]
\centering
\includegraphics[angle=0, width=0.98\textwidth]{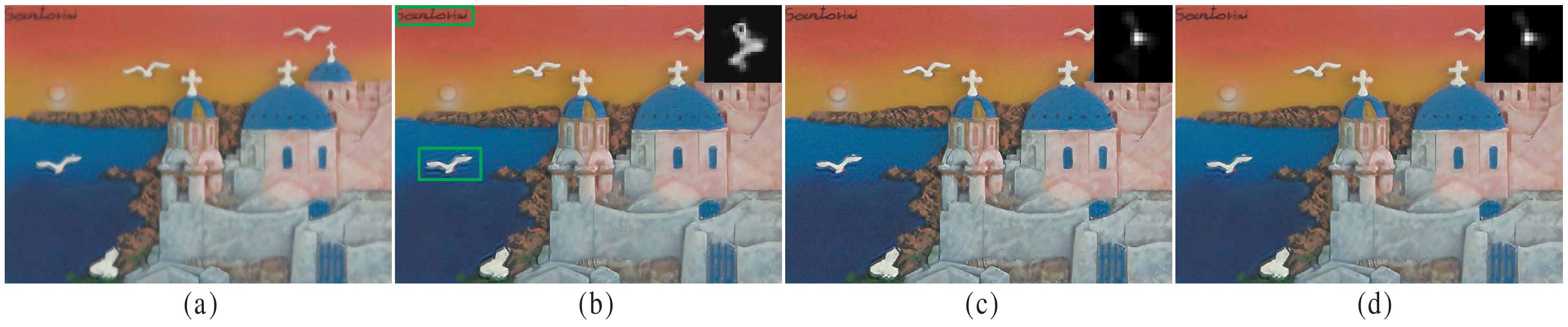}
\caption{Blurred image with some noise.
(a) Blurred image.
(b) Result of~\cite{Tai/noise/cvpr12}.
(c) Our result.
(d) The deblurred result is obtained by method~\cite{Cho/iccv11},
but the kernel is obtained by our method.
The green boxes shown in (b) contain some ringing artifacts or blur.
}
\label{fig: noise_image}
\end{figure}

From the results shown in Fig.~\ref{fig: noise_image},
one can see that our estimated results are comparable with that of~\cite{Tai/noise/cvpr12}.
%Compared with Figs.~(\ref{fig: noise_image})(b) and (d),
%one can see that our deblurred result is better than that of~\cite{Tai/noise/cvpr12}.

\begin{figure}[h]
\centering
\includegraphics[angle=0, width=0.98\textwidth]{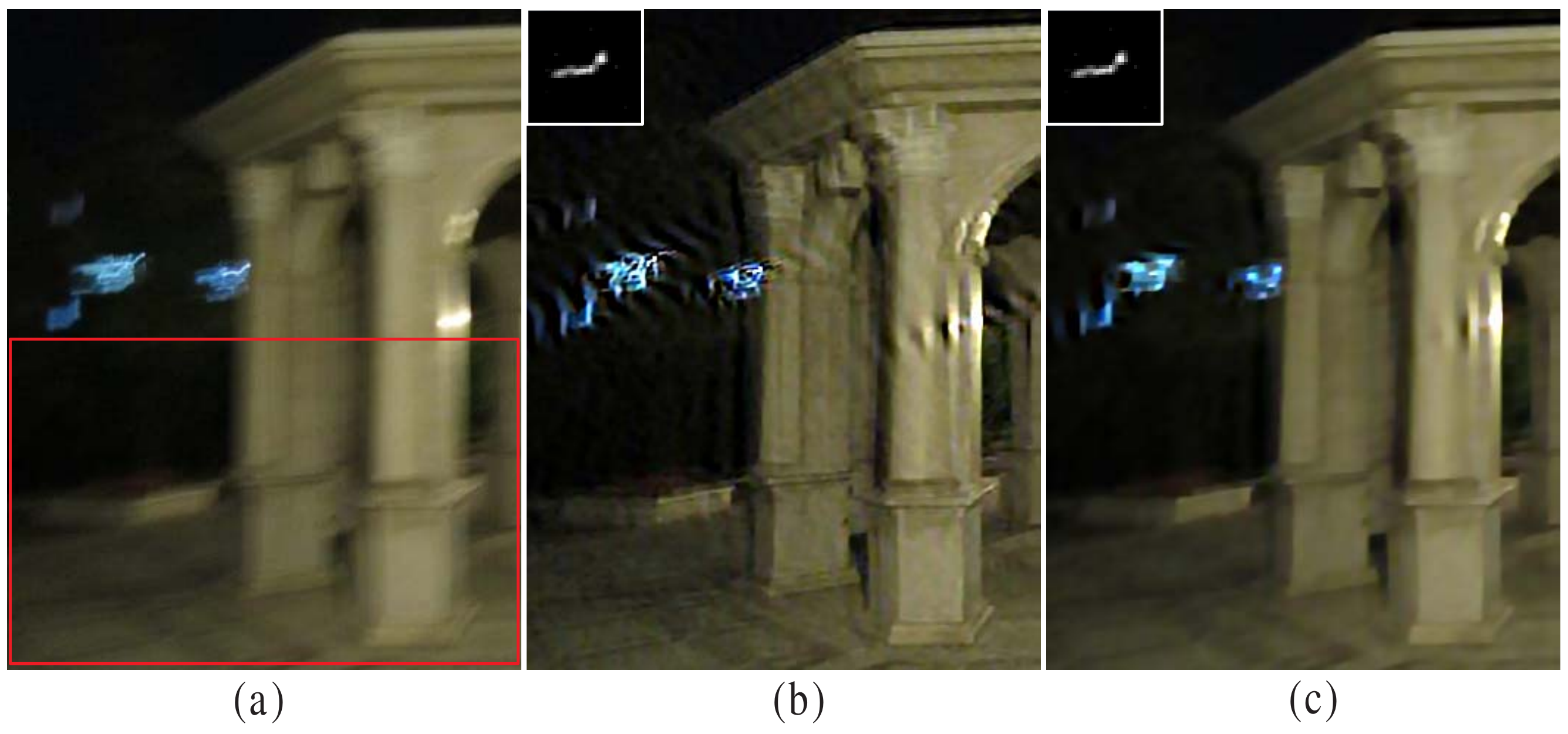}
\caption{Blurred image with some saturated areas.
(a) Blurred image.
(b) Our result.
(c) The deblurred result is obtained by method~\cite{Cho/iccv11},
but the kernel is obtained by our method.
The part shown in the red box in (a) is used to estimate the blur kernel.
}
\label{fig: our-captured}
\end{figure}

Fig.~\ref{fig: our-captured}(a) shows a real blurred image with some saturated areas.
Like the strategies~\cite{Cho/iccv11,Whyte/iccv2011/workshop},
we crop a rectangular region without obvious saturated pixels from Fig.~\ref{fig: our-captured}(a)
and estimate the blur kernel using the rectangular region (i.e., the part in
the red box shown in Fig.~\ref{fig: our-captured}(a)).
We then use method~\cite{Cho/iccv11} to restore the latent image.

One can see that our method provides a reliable kernel.
Due to the influence of saturated areas, the restored image shown in
Fig.~\ref{fig: our-captured}(b) contains some visual ringing artifacts.

These two examples, Figs.~\ref{fig: noise_image} and~\ref{fig: our-captured},
further demonstrate the effectiveness of our kernel estimation method.
However, it is noted that model~(\ref{eq:final image deblurring}) is not
robust to outliers. The deblurred results shown in Figs.~\ref{fig: noise_image}(c)
and~\ref{fig: our-captured}(b) demonstrate its limitations.
Thus, developing a better non-blind method will be an interesting work.

\section{Conclusion and Discussion}
In this work, we developed a novel kernel estimation algorithm based
on image salient edges. We discovered that image details could
undermine the kernel estimation, especially for large blur kernels.
Therefore, we proposed a self-adaptive algorithm which is able to remove
structures with potential aversive effects to the estimation. Our
kernel estimation model is able to remove noise and preserve the
characteristics of the kernel, such as continuity and sparsity, which
further reduces the
aversive effects caused by the wrong chosen structures. In the final
deconvolution step, we utilized the structural information for an
adaptive weighted regularization term to guide the latent image
restoration, which preserves the image details well.

We have extensively tested our algorithm, and found that it is able
to deblur images with both small and large blur kernels especially
when the blurred images contain rich details.

Our kernel estimation method would fail when the blurred image is textureless or contains severe saturated pixels.
If the blurred image is textureless, we will not obtain salient edges for kernel estimation.
If the blurred image contains a lot of saturated areas which distribute uniformly in the blurred image,
these saturated areas will be chosen for kernel estimation due to their saliency.
Since saturation breaks the linearity of the convolution-based blur model~(\ref{blur_model}),
this will inevitably damage the kernel estimation.
In addition, a spatially varying blur would not be properly handled by our method.
We leave these problems as our future work.

 \appendix

 \section{Relationship to the Structure Extraction Method~\cite{Xu/rtv}}
 \label{Connections to the Structure Extraction Method}
 The work in~\cite{Xu/rtv} used local information to accomplish texture removal
 and they proposed a new adaptive regularization term named relative
 total variation (RTV) which is defined as
\begin{eqnarray}\label{eq:rtv}
\mathfrak{R}_x(\textbf{x})=\frac{\sum_{\textbf{y}\in N_h(\textbf{x})}g_{\textbf{x},\textbf{y}}|\partial_x S(\textbf{y})|}{|\sum_{\textbf{y}\in N_h(\textbf{x})}g_{\textbf{x},\textbf{y}}\partial_x S(\textbf{y})| + \epsilon},\nonumber\\
\mathfrak{R}_y(\textbf{x})=\frac{\sum_{\textbf{y}\in N_h(\textbf{x})}g_{\textbf{x},\textbf{y}}|\partial_y S(\textbf{y})|}{|\sum_{\textbf{y}\in N_h(\textbf{x})}g_{\textbf{x},\textbf{y}}\partial_y S(\textbf{y})| + \epsilon},
\end{eqnarray}
where $S$ is the structure that we want to get
and $g_{\textbf{x},\textbf{y}}$ is a weighting function defined according to spatial affinity.
If $g_{\textbf{x},\textbf{y}}$ is a scalar weight, then the effect of Eq.~(\ref{eq:rtv}) is
similar to that of $1/r(\textbf{x})$ or $w(\textbf{x})$.
In fact, we can also use the variation form of Eq.~(\ref{eq:r-map}), i.e.,
\begin{eqnarray}\label{eq:rmap-appdenix}
\mathfrak{R}(\textbf{x}) = \frac{\sum_{\textbf{y}\in N_h(\textbf{x})}\|\nabla S(\textbf{y})\|}{\|\sum_{\textbf{y}\in N_h(\textbf{x})}\nabla S(\textbf{y})\| + \epsilon}, \nonumber
\end{eqnarray}
as a special regularizer to extract structures from an image.
However, our structure extraction method is different from~\cite{Xu/rtv}.
We use $w(\textbf{x})$ as an adaptive smoothness weight.
%-that is, we do not use Eq.~(\ref{eq:r-map}) as the regularizer.
Regarding the effect of RTV, we believe that it would extract some useful structures for kernel estimation.

\section*{Acknowledgements}
This work is supported by the
Natural Science Foundation of China-Guangdong Joint Fund under Grant
No.~U0935004, the Natural Science Foundation of China under Grant Nos. 61173103 and 91230103,
National Science and Technology Major Project under Grant No. 2013ZX04005021,
and China Postdoctoral Science Foundation under Grant No. 2013M530917.
The authors would like to thank Prof. Zhouchen Lin at Peking University
for his valuable comments.
Jinshan Pan would like to thank Dr. Li Xu at The Chinese University
of Hong Kong for some helpful discussions
and Dr. Sunghyun Cho at Adobe Research for providing his executable program of~\cite{Cho/et/al}.

%% Reference

\bibliographystyle{elsarticle-num}
\bibliography{egbib}

\begin{thebibliography}{10}
\expandafter\ifx\csname url\endcsname\relax
  \def\url#1{\texttt{#1}}\fi
\expandafter\ifx\csname urlprefix\endcsname\relax\def\urlprefix{URL }\fi
\expandafter\ifx\csname href\endcsname\relax
  \def\href#1#2{#2} \def\path#1{#1}\fi

\bibitem{Cho/et/al}
S.~Cho, S.~Lee, Fast motion deblurring, ACM Transactions on Graphics (SIGGRAPH
  Asia) 28~(5) (2009) 145.

\bibitem{Shan/et/al}
Q.~Shan, J.~Jia, A.~Agarwala, High-quality motion deblurring from a single
  image, ACM Transactions on Graphics 27~(3) (2008) 73.

\bibitem{Xu/et/al}
L.~Xu, J.~Jia, Two-phase kernel estimation for robust motion deblurring, in:
  ECCV, 2010, pp. 157--170.

\bibitem{Fergus/et/al}
R.~Fergus, B.~Singh, A.~Hertzmann, S.~T. Roweis, W.~T. Freeman, Removing camera
  shake from a single photograph, ACM Transactions on Graphics 25~(3) (2006)
  787--794.

\bibitem{Joshi/et/al}
N.~Joshi, R.~Szeliski, D.~J. Kriegman, {PSF} estimation using sharp edge
  prediction, in: CVPR, 2008, pp. 1--8.

\bibitem{Krishnan/CVPR2011}
D.~Krishnan, T.~Tay, R.~Fergus, Blind deconvolution using a normalized sparsity
  measure, in: CVPR, 2011, pp. 2657--2664.

\bibitem{Levin/CVPR2011}
A.~Levin, Y.~Weiss, F.~Durand, W.~T. Freeman, Efficient marginal likelihood
  optimization in blind deconvolution, in: CVPR, 2011, pp. 2657--2664.

\bibitem{Chen/et/al}
W.~G. Chen, N.~Nandhakumar, W.~N. Martin, Image motion estimation from motion
  smear-a new computational model, IEEE Transactions on Pattern Analysis
  Machine Intelligence 18~(1) (1996) 234--778.

\bibitem{Chan/and/Wong}
T.~Chan, C.~Wong, Total variation blind deconvolution, IEEE Transactions on
  Image Processing 7~(3) (1998) 370--375.

\bibitem{Cai/cvpr09}
J.-F. Cai, H.~Ji, C.~Liu, Z.~Shen, Blind motion deblurring from a single image
  using sparse approximation, in: CVPR, 2009, pp. 104--111.

\bibitem{Levin/CVPR2009}
A.~Levin, Y.~Weiss, F.~Durand, W.~T. Freeman, Understanding and evaluating
  blind deconvolution algorithms, in: CVPR, 2009, pp. 1964--1971.

\bibitem{Goldstein/eccv2012}
A.~Goldstein, R.~Fattal, Blur-kernel estimation from spectral irregularities,
  in: ECCV, 2012, pp. 622--635.

\bibitem{Joshi/phd}
N.~Joshi, Enhancing photographs using content-specific image priors, Ph.D.
  thesis, University of California (2008).

\bibitem{radon/cvpr/ChoPHF11}
T.~S. Cho, S.~Paris, B.~K.~P. Horn, W.~T. Freeman, Blur kernel estimation using
  the radon transform, in: CVPR, 2011, pp. 241--248.

\bibitem{ISD}
Y.~Wang, W.~Yin, Compressed sensing via iterative support detection, Tech.
  rep., Rice CAAM Technical Report TR09-30 (2009).

\bibitem{hu/eccv12/region}
Z.~Hu, M.-H. Yang, Good regions to deblur, in: ECCV, 2012, pp. 59--72.

\bibitem{Lucy/deconvolution}
L.~B. Lucy, An iterative technique for the rectification of observed
  distributions, Astronomy Journal 79~(6) (1974) 745--754.

\bibitem{Yuan/non/blind/deblurring/tog08}
L.~Yuan, J.~Sun, L.~Quan, H.-Y. Shum, Progressive inter-scale and intra-scale
  non-blind image deconvolution, ACM Transactions on Graphics 27~(3) (2008) 74.

\bibitem{Levin/et/al}
A.~Levin, R.~Fergus, F.~Durand, W.~T. Freeman, Image and depth from a
  conventional camera with a coded aperture, ACM Transactions on Graphics
  26~(3) (2007) 70--78.

\bibitem{joshi/color/prior/cvpr09}
N.~Joshi, C.~L. Zitnick, R.~Szeliski, D.~J. Kriegman, Image deblurring and
  denoising using color priors, in: CVPR, 2009, pp. 1550--1557.

\bibitem{Wang/Yang/Yin/Zhang}
Y.~Wang, J.~Yang, W.~Yin, Y.~Zhang, A new alternating minimization algorithm
  for total variation image reconstruction, SIAM Journal on Imaging Sciences
  1~(3) (2008) 248--272.

\bibitem{Whyte/cvpr10}
O.~Whyte, J.~Sivic, A.~Zisserman, J.~Ponce, Non-uniform deblurring for shaken
  images, in: CVPR, 2010, pp. 491--498.

\bibitem{non/uniform/deblur/joshi}
N.~Joshi, S.~B. Kang, C.~L. Zitnick, R.~Szeliski, Image deblurring using
  inertial measurement sensors, ACM Transactions on Graphics 29~(4) (2010) 30.

\bibitem{Gupta/non/uniform/eccv10}
A.~Gupta, N.~Joshi, C.~L. Zitnick, M.~F. Cohen, B.~Curless, Single image
  deblurring using motion density functions, in: ECCV, 2010, pp. 171--184.

\bibitem{hirsch/iccv11/non/uniform/deblurring}
M.~Hirsch, C.~J. Schuler, S.~Harmeling, B.~Sch{\"o}lkopf, Fast removal of
  non-uniform camera shake, in: ICCV, 2011, pp. 463--470.

\bibitem{hui/ji/cvpr12/non/uniform/deblurring}
H.~Ji, K.~Wang, A two-stage approach to blind spatially-varying motion
  deblurring, in: CVPR, 2012, pp. 73--80.

\bibitem{Rudin/Osher/Fatemi}
L.~I. Rudin, S.~Osher, E.~Fatemi, Nonlinear total variation based noise removal
  algorithms, Physica D 60 (1992) 259--268.

\bibitem{Shock/filter}
S.~Osher, L.~I. Rudin, Feature-oriented image enhancement using shock filters,
  SIAM Journal on Numerical Analysis 27~(4) (1990) 919--940.

\bibitem{Chen/Jia}
J.~Chen, L.~Yuan, C.~K. Tang, L.~Quan, Robust dual motion deblurring, in: CVPR,
  2008, pp. 1--8.

\bibitem{Xu/L0/smooth}
L.~Xu, C.~Lu, Y.~Xu, J.~Jia, Image smoothing via {\it l}$_{\mbox{0}}$ gradient
  minimization, ACM Transactions on Graphics (SIGGRAPH Asia) 30~(6) (2011) 174.

\bibitem{edge/preserving/tog/08}
Z.~Farbman, R.~Fattal, D.~Lischinski, R.~Szeliski, Edge-preserving
  decompositions for multi-scale tone and detail manipulation, ACM Transactions
  on Graphics 27~(3) (2008) 67.

\bibitem{huiji/tip/JiW12}
H.~Ji, K.~Wang, Robust image deblurring with an inaccurate blur kernel, IEEE
  Transactions on Image Processing 21~(4) (2012) 1624--1634.

\bibitem{Cho/iccv11}
S.~Cho, J.~Wang, S.~Lee, Handling outliers in non-blind image deconvolution,
  in: ICCV, 2011, pp. 495--502.

\bibitem{Whyte/iccv2011/workshop}
O.~Whyte, J.~Sivic, A.~Zisserman, Deblurring shaken and partially saturated
  images, in: ICCV Workshops, 2011, pp. 745--752.

\bibitem{Tai/noise/cvpr12}
Y.-W. Tai, S.~Lin, Motion-aware noise filtering for deblurring of noisy and
  blurry images, in: CVPR, 2012, pp. 17--24.

\bibitem{Xu/rtv}
L.~Xu, Q.~Yan, Y.~Xia, J.~Jia, Structure extraction from texture via relative
  total variation, ACM Transactions on Graphics (SIGGRAPH Asia) 31~(6) (2012)
  139.

\end{thebibliography}

\end{document}